\documentclass[runningheads]{llncs}

\usepackage{eccv}

\usepackage{eccvabbrv}

\usepackage{graphicx}
\usepackage{booktabs}
\usepackage{nicematrix}
\usepackage{multirow, makecell}
\usepackage{threeparttable}
\usepackage[symbol]{footmisc}

\usepackage[accsupp]{axessibility}  
\usepackage{xcolor,colortbl}

\usepackage{hyperref}

\usepackage{amssymb}
\usepackage{pifont}
\usepackage{orcidlink}
\usepackage{kotex}
\usepackage{algpseudocode}
\usepackage{algorithm}

\definecolor{LightCyan}{rgb}{0.88,1,1}
\newcommand{\xmark}{\ding{55}}%

\begin{document}

\title{Efficient Diffusion-Driven Corruption Editor \\for Test-Time Adaptation}

\titlerunning{Decorruptor}

\author{Yeongtak Oh$^*$\inst{1}\orcidlink{0000-0002-3642-1830}, ~~
Jonghyun Lee$^*$\inst{1}\orcidlink{0000-0002-1530-1020}, ~~
Jooyoung Choi\inst{1}\orcidlink{0009-0009-3862-0639}, ~~
Dahuin Jung\inst{3}\orcidlink{0000-0002-1344-1054}, ~~
Uiwon Hwang$^{\dagger}$\inst{4}\orcidlink{0000-0001-5054-2236}, ~~
Sungroh Yoon$^{\dagger}$\inst{1}\inst{2}\orcidlink{0000-0002-2367-197X}}

\authorrunning{Oh. and Lee. et al.}
\institute{Department of Electrical and Computer Engineering, Seoul National University \and Interdisciplinary Program in Artificial Intelligence, Seoul National University \and 
School of Computer Science and Engineering, Soongsil University \and
Division of Digital Healthcare, Yonsei University \\ 
\email{dualism9306@snu.ac.kr, leejh9611@snu.ac.kr, jy\_choi@snu.ac.kr, dahuin.jung@ssu.ac.kr, uiwon.hwang@yonsei.ac.kr, sryoon@snu.ac.kr}}

\def\thefootnote{*}\footnotetext{These authors contributed equally to this work}
\def\thefootnote{$\dagger$}\footnotetext{Corresponding authors}

\authorrunning{Decorruptor} 


\maketitle

\renewcommand{\thefootnote}{\fnsymbol{footnote}}

\begin{abstract}
Test-time adaptation (TTA) addresses the unforeseen distribution shifts occurring during test time. In TTA, performance, memory consumption, and time consumption are crucial considerations.
A recent diffusion-based TTA approach for restoring corrupted images involves image-level updates. However, using pixel space diffusion significantly increases resource requirements compared to conventional model updating TTA approaches, revealing limitations as a TTA method. To address this, we propose a novel TTA method that leverages an image editing model based on a latent diffusion model (LDM) and fine-tunes it using our newly introduced corruption modeling scheme. This scheme enhances the robustness of the diffusion model against distribution shifts by creating (clean, corrupted) image pairs and fine-tuning the model to edit corrupted images into clean ones. Moreover, we introduce a distilled variant to accelerate the model for corruption editing using only 4 network function evaluations (NFEs). We extensively validated our method across various architectures and datasets including image and video domains. Our model achieves the best performance with a 100 times faster runtime than that of a diffusion-based baseline. Furthermore, it is three times faster than the previous model updating TTA method that utilizes data augmentation, making an image-level updating approach more feasible.
\footnote[1]{Project page: \textcolor{magenta}{\href{https://github.com/oyt9306/Decorruptor}{https://github.com/oyt9306/Decorruptor}}}

\end{abstract}
\section{Introduction}
Test-time adaptation (TTA)~\cite{tent} is a task aimed at achieving higher performance than simple inference when there is a distribution shift between source and target domain, using a minimal resource (\eg, inference time and memory consumption) overhead. Traditional TTA methodologies~\cite{tent, sar, eata, boudiaf2022parameter} primarily update only a subset of model parameters or manipulate the model's output to obtain predictions adapted to the target distribution. However, these methodologies show sensitive performance under wild scenarios~\cite{sar, deyo} (\eg, biased, label shifts, mixed, and batch size $1$ scenarios) and episodic setting~\cite{cotta}.

Gao~\etal.~\cite{dda} first proposed a diffusion-based image-level (input) updating approach for TTA called diffusion-driven adaptation (DDA), which restores the input image via an ImageNet~\cite{deng2009imagenet} pre-trained pixel-space diffusion model \cite{dhariwal2021diffusion}. DDA shows more robust performance than model-updating TTA approaches~\cite{tent, zhang2022memo} under episodic settings and consistent performance enhancement with various architectures. However, the backbone diffusion model, DDPM~\cite{ho2020denoising}, requires large memory consumption and a significant amount of inference time. Given the resource and time constraints inherent in TTA, implementing DDA in real-world applications is not feasible. Therefore, for the effective usability of an image-level updating TTA, it is essential to incorporate fast and lightweight input updates.
\begin{figure}[t!]
    \centering
    \includegraphics[width=0.75\linewidth]{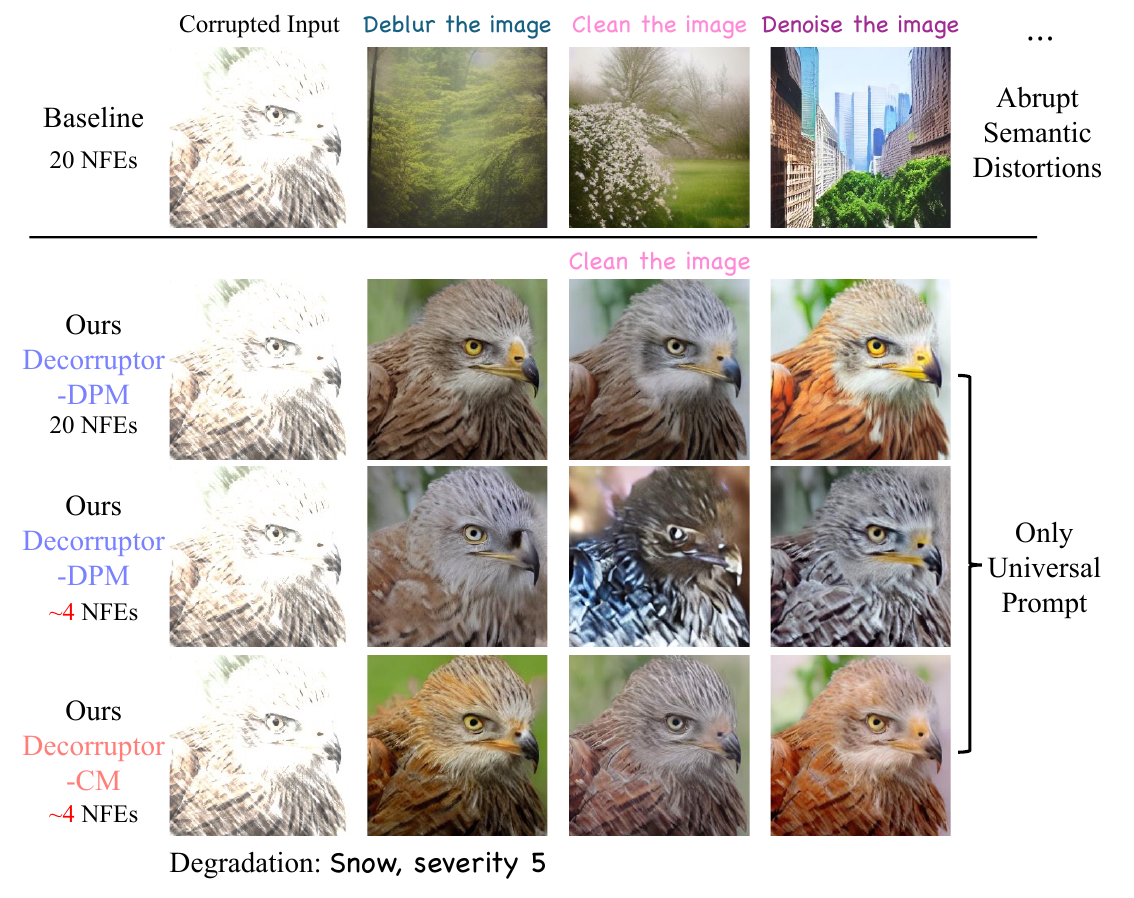} 
    \caption{Visualization of instruction-guided image editing for the unseen corrupted image at the test-time. Compared to the baseline IP2P method, our proposed Decorruptor-DPM ($20$-step) and Decorruptor-CM ($4$-step) models show effective editing results without hurting the original semantics of the input corrupted image.}
    \label{fig:motivation}
\end{figure}

To achieve efficient adaptation, we consider using Instruct-Pix2Pix (IP2P)~\cite{brooks2023instructpix2pix}, a method that edits images in the compressed latent space. IP2P takes both text instructions and images as conditions to enable instruction-based image editing. This approach leverages not pixel space but latent space, which facilitates the efficient generation of images. However, as shown in the first row of Fig.~\ref{fig:motivation}, IP2P model is infeasible for TTA scenarios where out-of-domain (\ie, corrupted) images become inputs, as it can only edit in-domain images into in-domain images. Thus, to utilize IP2P model in TTA scenarios, it is crucial to enhance its robustness under distribution shifts, including test-time corruption.

In this paper, we propose a new diffusion-based input updating TTA methodology named Decorruptor using diffusion probabilistic model (Decorruptor-DPM) that can efficiently respond to unseen corruptions. To enhance the diffusion model’s robustness, we draw inspiration from data augmentation methods~\cite{cubuk2019autoaugment, zhang2017mixup, yun2019cutmix, hendrycks2022pixmix}, known for their efficacy in enhancing robustness against distribution shifts. In response, Decorruptor-DPM applies a novel \textit{corruption modeling} scheme to IP2P: generate (clean, corrupted) image pairs and use them for fine-tuning to facilitate the restoration of corrupted images to their clean counterparts. 
To the best of our knowledge, the application of data augmentation for enhancing the robustness of the diffusion model appears to be unexplored in existing literature. Decorruptor-DPM supports efficient editing against test-time corruption as can be seen in Fig.~\ref{fig:motivation}, requiring only a universal instruction. In addition, Decorruptor-DPM enables \underline{46 times faster} input updates than DDA owing to the latent-level computation and fewer generation steps. 
To be practically applicable with TTA, where inference time is crucial, we further propose Decorruptor using consistency model (Decorruptor-CM), the accelerated variant of Decorruptor-DPM, by distilling the diffusion using consistency distillation~\cite{luo2023latent}. As shown in Fig.~\ref{fig:motivation}, Decorruptor-CM achieves similar corruption editing effects to Decorruptor-DPM’s 20 network function evaluations (NFEs) with only 4 NFEs. 

We assess the performance of data edited by our models on the ImageNet-C~\cite{imagenet-c} and ImageNet-$\bar{\mathrm{C}}$~\cite{imagenet-c-bar}, with various architectures. With around \underline{100 times} \underline{faster runtime}, our models exhibited the best performance on ImageNet-C and -$\bar{\mathrm{C}}$ across various architectures. 
Notably, Decorruptor-CM demonstrated superior performance compared to MEMO~\cite{zhang2022memo}, an image augmentation-based model updating TTA method, achieving three times speed enhancements.
Contrary to DDA, which shows a performance drop across certain architectures, our approach reveals improvements in all evaluated architectures. Additionally, by harnessing its rapid inference, Decorruptor-CM extends its applicability from the image to the video domain, showcasing outperforming editing outcomes on the UCF-101~\cite{soomro2012ucf101} video dataset when compared to DDA.
Our contributions are as follows:
\begin{itemize}
    \item We propose Decorruptor-DPM that enhances the \textcolor{black}{robustness} and \textcolor{black}{efficiency} of diffusion-based input updating approach for TTA through the incorporation of a novel \textcolor{black}{\textit{corruption modeling}} scheme within the \textcolor{black}{LDM}.
    \item We propose Decorruptor-CM, as an accelerated model, by distilling the DPM to significantly reduce inference time with minimal performance degradation. By ensembling multiple edited images' predictions, Decorruptor-CM even achieves higher classification accuracy than Decorruptor-DPM while being faster in execution.
    \item We demonstrate high performance and generalization capabilities with a faster runtime through extensive experiments on image and video TTA. Decorruptor-CM shows three times faster runtime than MEMO, making an input updating approach more practical.
\end{itemize}

\section{Related Works}
\subsection{Latent Diffusion Models}
The latent diffusion model (LDM)~\cite{rombach2022high} is a representative method that overcomes the large memory/time consumption drawbacks of DDPM~\cite{ho2020denoising}. LDM reduces memory consumption and inference time by performing the denoising process in latent space instead of pixel space. Stable diffusion (SD)~\cite{rombach2022high}, a scaled-up version of LDM, is a large-scale pre-trained text-to-image diffusion model that has shown unprecedented success in high-quality and diverse image synthesis. Unlike previous SD-based image editing methodologies~\cite{hertz2022prompt, parmar2023zero} which require paired texts in the image editing stage, InstructPix2Pix~\cite{brooks2023instructpix2pix} enables image editing solely based on instructions. However, as IP2P only supports clean input images, we enable corruption editing by fine-tuning the diffusion models with our proposed corruption modeling scheme. 

\subsection{Image Restoration} 
In the image restoration (IR) task, several works have been proposed to exploit the advantages of diffusion models. To solve linear inverse problems for IR tasks such as inpainting, denoising, deblurring, and super-resolution, applying SD~\cite{ai2023multimodal, chung2023prompt, jiang2023autodir} in image restoration has recently emerged. However, since SD relies on classifier-free guidance~\cite{ho2022classifier} with text-conditioning for image editing, a significant limitation exists to applying SD itself to TTA to remove arbitrary unseen test-time corruptions. To be specific, it is infeasible as previous text-guided image editing methods used for image restoration domain require prior knowledge of text information corresponding to the test-time corruption~\cite{ai2023multimodal, chung2023prompt, jiang2023autodir}, or necessitates such as a blur kernel or other degradation matrices~\cite{chung2023diffusion, wang2022zero, kawar2022denoising}. In this paper, we elucidate that our work is significantly different from IR tasks, as we do not require either pre-defined corruption information or degradation matrices for corruption editing at test time. 
\begin{table*}[t!]
    \caption{Comparisons with multiple image-to-image tasks. IN indicates ImageNet. Our Decorruptor shows efficiency, generalizability, and high performance.}
    \label{tab:imagenet-c-normal}
    \newcolumntype{a}{>{\columncolor{cyan!10}}c}
    \vspace{-0.1in}

    \setlength{\tabcolsep}{0.5em} 

    \begin{center}
        \begin{threeparttable}
            \LARGE
            \resizebox{1.0\linewidth}{!}{
                \begin{tabular}{llcccccc}
                    \toprule
                    \multirow{2}{*}{TTA requirements} & & Image Editing & \multicolumn{2}{c}{Image Reconstruction} & \multicolumn{2}{c}{Image Decorruption} \\
                    \cmidrule(lr){3-3}\cmidrule(lr){4-5}\cmidrule(lr){6-7}

                    & & InstructPix2Pix~\cite{brooks2023instructpix2pix} & DDRM~\cite{ddrm} & DPS~\cite{chung2023diffusion} & {DDA~\cite{dda}} & \cellcolor{cyan!10}Ours (DPM / CM) \\

                    \midrule

                    \multirow{2}{*}{\begin{tabular}[c]{@{}l@{}}Efficiency\\ (Minimal overhead)\end{tabular}} & NFEs & 20 & 20 & 1000 & 50 & \cellcolor{cyan!10}20 / 4 \\
                    & Noise space & Latent space & Pixel space & Pixel space & Pixel space & \cellcolor{cyan!10}Latent space \\
                    \midrule

                    {Generalization} & Degradation type & \textcolor{red}{\xmark} & Pre-defined & Pre-defined & Unseen & \cellcolor{cyan!10}Unseen \\
                    \midrule

                    \multirow{2}{*}{Performance} & IN-C Acc (\%) & \textcolor{red}{\xmark} & \textcolor{red}{\xmark} & \textcolor{red}{\xmark} & 29.7 & 30.5 / \cellcolor{cyan!10}\textbf{32.8} \\
                    & IN-$\bar{\mathrm{C}}$ Acc (\%) & \textcolor{red}{\xmark} & \textcolor{red}{\xmark} & \textcolor{red}{\xmark} & 29.4 & 41.8 / \cellcolor{cyan!10}\textbf{47.1} \\
                    \bottomrule
                \end{tabular} 
            }
        \end{threeparttable}
    \end{center}
    \vspace{-0.25in}
\end{table*}

\subsection{Test-Time Adaptation}
TTA aims to enhance inference performance with minimal resource overhead under distribution shifts. In contrast to unsupervised domain adaptation (UDA)~\cite{uda1, uda2, uda3}, TTA lacks access to the source data. Moreover, unlike source-free domain adaptation~\cite{shot, cowa, sfdada}, TTA has an online characteristic, obtaining target data only once through streaming. A prominent approach in TTA involves updating only a subset of parameters~\cite{tent, eata, sar, deyo, niu2024test}. However, model updating TTA methods face a risk of catastrophic forgetting~\cite{eata, cotta} as it lacks access to source data during training. Furthermore, the absence of clear criteria for hyperparameter selection poses a drawback, making it challenging to ensure performance in practical applications~\cite{TTA_hp}.
Gao \etal~\cite{dda} presents diffusion-driven adaptation (DDA) to overcome these limitations. DDA utilizes a pre-trained diffusion model to transform corrupted input images into clean in-distribution images, updating the input images instead of the model. This approach enhances robustness in single-image evaluation as well as in ordered data scenarios. However, DDA falls short of meeting the efficiency requirement of TTA, as obtaining predictions for a single sample takes a long time. We greatly overcome such efficiency drawbacks. By combining LDM and CM, our model achieves higher performance and significantly reduced inference time compared to DDA.
The overview of comparisons with related works is summarized in Table~\ref{tab:imagenet-c-normal}.
\begin{figure}[t!]
    \centering
    \includegraphics[width=\linewidth]{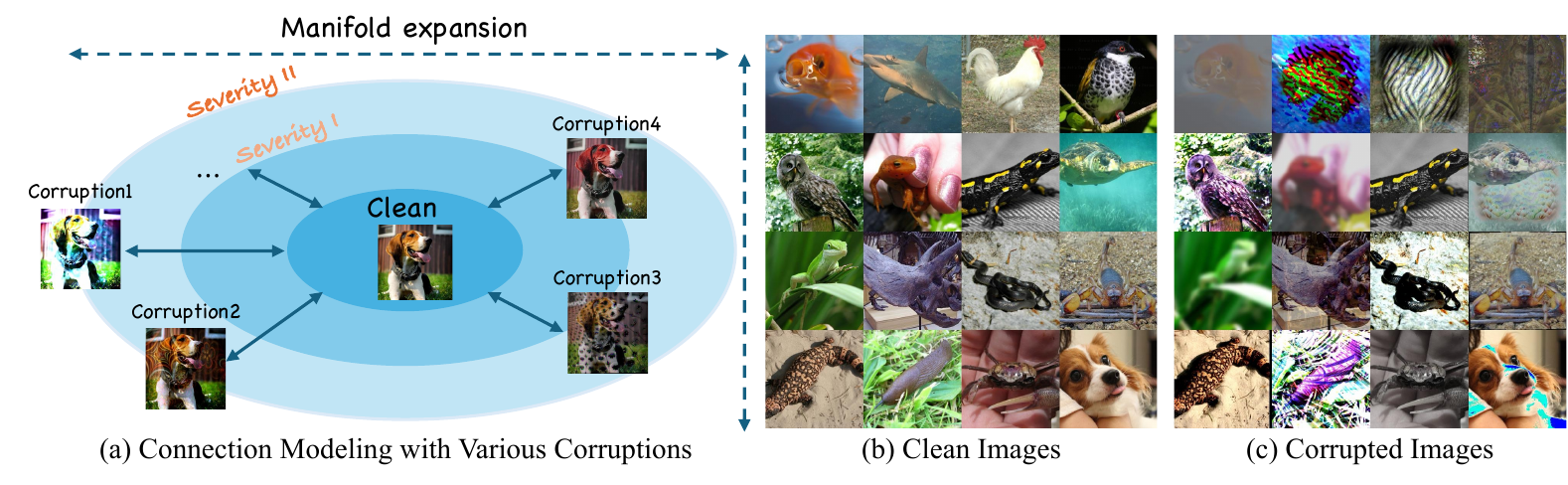}
    \caption{Representations of (a) Instance-wise connection map for corruption-like augmentations, corruption crafting results of (b) clean images to (c) corrupted images. In (a), we showcase how we constitute various corruption-like augmentations. Here, the sensitivity means the granularity of the corruption, the crafting phase means how to create the corrupted images, and the learning phase means how to learn editing. 
    }
    \label{fig:connection_map}
\end{figure}
\section{Preliminaries}
\subsection{Diffusion Models}
Diffusion models~\cite{sohl2015deep,ho2020denoising}, also known as score-based generative models~\cite{song2019generative,song2020score}, are a popular family of generative models that generate data from Gaussian noise. Specifically, these models learn to reverse a diffusion process that translates the original data distribution \( p_{\text{data}}(x) \) towards a marginal distribution \( q_t(x_t) \), facilitated by a transition kernel defined as \( q_{0t}(x_t | x_0) = \mathcal{N}(x_t | \alpha(t)x_0, \sigma^2(t)I) \), in which \( \alpha(t) \) and \( \sigma(t) \) are pre-defined noise schedules. Viewed from a continuous-time perspective, this diffusion process can be modeled by a stochastic differential equation (SDE)~\cite{song2020score, karras2022elucidating} over the time interval \([0, T]\). To learn the reverse of SDE, diffusion models are trained to estimate score function $\epsilon_\theta(z_t,t)$ with U-Net architecture~\cite{ronneberger2015u}.
Song \etal~\cite{song2020score} show that the reverse process of SDE has its corresponding \textit{probability flow original differential equation} (PF-ODE).

\subsubsection{Classifier Free Guidance}
During inference, diffusion models use the classifier-free guidance  (CFG)~\cite{ho2022classifier} to ensure the input conditions such as text and class labels.
Compared to the classifier guidance~\cite{dhariwal2021diffusion} that requires training an additional classifier, the CFG operates without the need for a pre-trained classifier. Instead, CFG utilizes a linear combination of score estimates from an unconditional diffusion model trained concurrently with a conditional diffusion model:
\begin{equation}
    \hat{\epsilon}_{\theta}(z_t, \omega, t, c) := (1 + \omega) \epsilon_{\theta}(z_t, t, c) - \omega \epsilon_{\theta}(z_t, t, \emptyset).
    \label{eq:cfg}
\end{equation}
\subsection{Consistency Models}\label{sec:cm}
Due to their iterative updating nature, the slow generation speed of diffusion models is a known limitation. To overcome this, the Consistency Model (CM)~\cite{song2023consistency} has been introduced as a new generative model that accelerates generation to a single step or a few steps. CM operates on the principle of mapping any point from the trajectory of the PF-ODE to its destination. This is achieved through a consistency function defined as \( f : (x_t, t) \mapsto x_{\varepsilon} \), where \( \varepsilon \) is a small positive value. A key aspect of CM is that the consistency function must fulfill the self-consistency property:
\begin{equation}
    f(x_t, t) = f(x_{t'}, t'), \quad \forall t, t' \in [\varepsilon, T].
\end{equation}
To ensure that $f_{\theta}(x, \varepsilon) = x$, the consistency model $f_{\theta}$ can be parameterized as $f_{\theta}(x, t) = c_{\text{skip}}(t)x + c_{\text{out}}(t)F_{\theta}(x, t)$, where $c_{\text{skip}}(t)$ and $c_{\text{out}}(t)$ are differentiable functions with $c_{\text{skip}}(\varepsilon) = 1$ and $c_{\text{out}}(\varepsilon) = 0$, and $F_{\theta}(x, t)$ is a deep neural network and it becomes $(\mathbf{z}_{t} - \sigma_t \hat{\epsilon}(\mathbf{z}, c, t))/ {\alpha_t}$ for $\epsilon$-prediction models like Stable Diffusion. One way to train a consistency model is to distill a pre-trained diffusion model, by training an online model $\theta$ while updating target model $\theta^-$ with exponential moving average (EMA), defined as $\theta^- \leftarrow \mu\theta^- + (1 - \mu)\theta$. The consistency distillation loss is defined as follows:
\begin{equation}
    \mathcal{L}(\theta, \theta^-; \Phi) = \mathbb{E}_{x_t} \left[ d \left( f_{\theta}(x_{t_{n}}, t_{n+1}), f_{\theta^-}\left(\hat{x}_{t_n}, t_n\right) \right) \right],
    \label{eq:cd_train}
\end{equation}
\noindent where $d(\cdot, \cdot)$ is a squared $\ell_2$ distance $d(x, y) = \|x - y\|^2$, and $\hat{x}_{t_n}$ is a one-step estimation of $x_{t_n}$ from $x_{t_{n+1}}$ as $\hat{x}_{t_n} \leftarrow x_{t_{n+1}} + (t_n - t_{n+1})\Phi(x_{t_{n+1}}, t_{n+1}; \Phi)$, where $\Phi$ denotes the numerical ODE solver like DDIM~\cite{song2020denoising}. 

Luo \etal \cite{luo2023latent} recently introduced Latent Consistency Models (LCM) which accelerate a text-to-image diffusion model. They propose a consistency function $f_{\theta} : (z_t, \omega, c, t) \mapsto z_0$ that directly predicts the solution of PF-ODE augmented by CFG, with additional guidance scale condition $\omega$ and text condition $c$. The LCM is trained by minimizing the loss
\begin{equation}
    \mathcal{L}_{LCD} (\theta, \theta^-; \Psi) = \mathbb{E}_{z_t, \omega, c, n} \left[ d \left( f_{\theta}(z_{t_{n+1}}, \omega, c, t_{n+1}), f_{\theta^-}(\hat{z}_{t_n}^{\Psi, \omega}, \omega, c, t_n) \right) \right].   
    \label{eq:LCD}
\end{equation}
Here, $\omega$ and $n$ are uniformly sampled from interval $[\omega_{\text{min}}, \omega_{\text{max}}]$ and $[1, \ldots, N-1]$ respectively. $\hat{z}_{t_n}^{\Psi, \omega}$ is estimated using the pre-trained diffusion model and PF-ODE solver $\Psi$~\cite{song2020denoising}, represented as follows:
\begin{equation}
\hat{z}_{t_n}^{\Psi, \omega} - z_{t_{n+1}} \approx (1 + \omega) \Psi(z_{t_{n+1}}, t_{n+1}, t_n, c) - \omega \Psi(z_{t_{n+1}}, t_{n+1}, t_n, \emptyset).
\label{eq_lcm}
\end{equation}
\begin{figure}[t!]
    \centering
    \includegraphics[width=\linewidth]{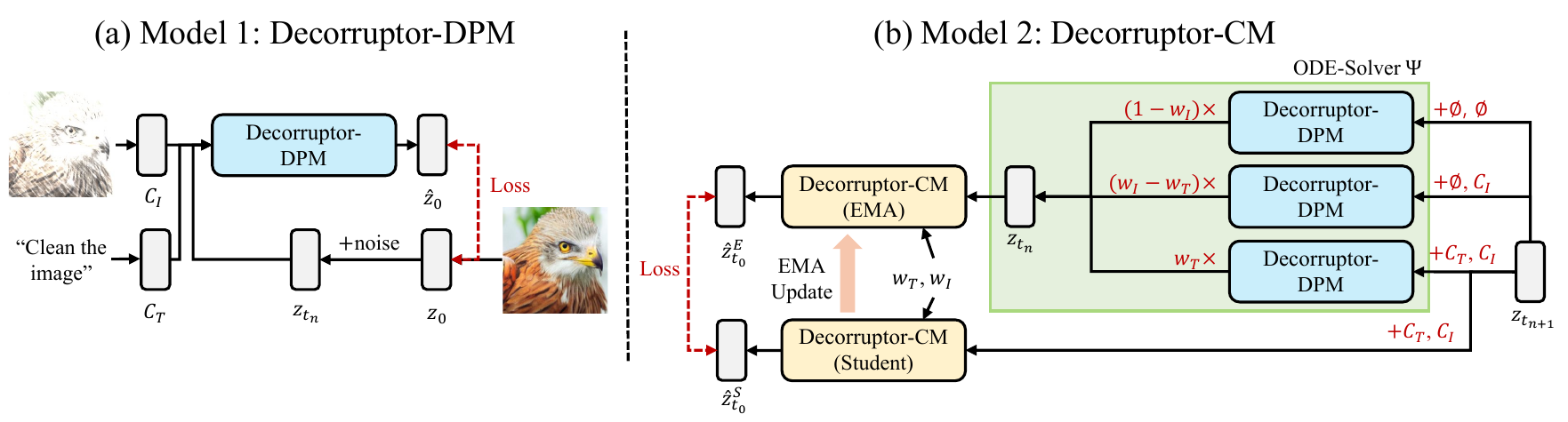} 
    \caption{Schematic of the overall training pipeline for the two proposed model variants: (a) Decorruptor-DPM, (b) Decorruptor-CM.}
    \label{fig:proposed_pipeline}
\end{figure}
\section{Proposed Method}
A crucial consideration of Decorruptor lies in the diversity of augmentations on corruption-like augmentations during the inductive learning stage. To this end, we elucidate how we get the paired data of clean and corrupted data in \ref{Section:corruption_modeling}, fine-tuning the models using those pairs in \ref{Section:DPM}, distilling the model for acceleration at inference time in \ref{Section:CM}, and explanations on overall process in \ref{Section:Overall}. Please refer to Appendix A.4 for the pseudo-codes of training and inference of Deccoruptor-CM.
\subsection{Corruption Modeling Scheme}
\label{Section:corruption_modeling}

As shown in the first row of Fig.~\ref{fig:motivation}, the pre-trained diffusion model does not generalize well to out-of-domain data not used during training. Therefore, to effectively utilize diffusion in TTA with incoming unseen corrupted data, the diffusion model also needs robustness against corruption. To this end, as a method to impose the diffusion model with robustness, we introduce a novel \textit{corruption modeling} scheme: create pairs of (clean, corrupted) images and utilize them for fine-tuning to enable the recovery of corrupted images to their clean states.
Through the training process of editing corrupted images onto clean ones, we broaden the diffusion model's manifold to edit corrupted inputs, enhancing robustness against unseen corruptions. 
To the best of our knowledge, this novel scheme to robustify the diffusion models has not been previously explored.

To create corrupted data, we employed prevalent data augmentation strategies on the given clean images. PIXMIX~\cite{hendrycks2022pixmix}, for example, performs data augmentation in an on-the-fly manner based on class-agnostic complex images. Furthermore, we also consider the widely-used data augmentation method for self-supervised contrastive learning, such as SimSiam~\cite{chen2021exploring}. To augment the image, for the crafting phase, the corrupted samples are easily crafted in a one-to-many manner from the clean image. In the following, for the learning phase, we utilized these corrupted samples for many-to-one training for corruption editing via our Decorruptor-DPM model with `\texttt{Clean the image}' instruction $c_T$. Thus, our model is trained to restore mixed complex corruptions from the diverse augmented samples $c_I$ onto clean image $x$. The connection modeling process is summarized in Fig.~\ref{fig:connection_map} (a).

\subsection{Decorruptor-DPM: Instruction-Based Corruption Editing}
\label{Section:DPM}
To train our diffusion model, as it diffuses the image in the latent space, the training objective can be rewritten as the following equation:
\begin{equation}
\mathcal{L}(\theta) = \mathbb{E}_{z\sim\mathcal{E}(x), c_T, \epsilon \sim \mathcal{N}(0,1), t} \left[ \| \epsilon - \epsilon_{\theta}(\mathbf{z}_t, t, c_T, c_I) \|^2 \right],
\label{eq:latent_loss}
\end{equation}
where $x$ denotes an image, $z$ is the encoded latent, $z_{t}$ denotes diffused latents at timestep $t\in T$, text condition $c_{T}$, and corrupted image condition $c_I$. 

\subsubsection{Fine-Tuning U-Net with Corruption-Like Augmentations}
We note that the large-scale LAION dataset \cite{schuhmann2022laion}, encompassing diverse image domains including art, 3D, and aesthetic categories, is highly different from the ImageNet dataset. Thus, exploiting the IP2P model itself, which is fine-tuned on generated samples with SD trained on the LAION dataset, inevitably generates domain-biased samples when using it for corruption editing of ImageNet data. 

To overcome such limitations, we have fine-tuned the diffusion model's U-Net~\cite{ronneberger2015u} initialized from the checkpoint of SD~\cite{rombach2022high} via IP2P training protocol on corrupted images using only ImageNet~\cite{deng2009imagenet} training data, ensuring they can edit images with the universal prompt alone. We use the same prompt used at training time that can revert unknown corruptions at inference time. This distinguishes our approach from previous works that required significant effort for prompt engineering. To facilitate image conditioning, following Brooks~\etal~\cite{brooks2023instructpix2pix}, we introduce extra input channels into the initial convolutional layer. The diffusion model's existing weights are initialized using pre-trained checkpoints, while the weights associated with the newly incorporated input channels are set to zero. The illustration of our Decorruptor-DPM is represented in Fig. \ref{fig:proposed_pipeline} (a). 

\subsubsection{Scheduling Image Guidance Scale}
At inference time, for Decorruptor-DPM, we utilized 20 DDIM~\cite{song2020denoising} steps and modified the image guidance scheduling to enable more effective editing. Unlike the text guidance scale, the image guidance scale is a hyper-parameter that determines how much of the input semantics are retained. Considering multi-modal conditioning guidances of the input image $c_{I}$ and text instructions $c_T$, CFG for our Decorruptor-DPM is as follows:
\begin{equation}
    \begin{aligned}
        \hat{\epsilon}_{\theta}(z_t, t, c_I, c_T) & = \epsilon_{\theta}(z_t, t, \emptyset, \emptyset) + \omega_I(t) (\epsilon_{\theta}(z_t, t, c_I, \emptyset) - \epsilon_{\theta}(z_t, t, \emptyset, \emptyset)) \\
        & + \omega_T(\epsilon_{\theta}(z_t, t, c_I, c_T) - \epsilon_{\theta}(z_t, t, c_I, \emptyset)).    
    \end{aligned}
    \label{eq:ip2p_mod}
\end{equation}
It is known that if the scale is too large, the image tends to remain almost unchanged during editing, and conversely, if it is too small, the original image semantics are ignored, relying solely on text guidance for editing \cite{brooks2023instructpix2pix}. Since our input image is corrupted, we notice that the large guidance scale is needed at the large timestep of the pure noise phase, and a smaller guidance scale is used at the near image phase. Thus, in Eq.~(\ref{eq:ip2p_mod}), we employed sqrt-scheduling for the image guidance scale $\omega_{I}$ by sampling it from $1.8$ to $0$ for $t\in [T,0]$. 

\subsection{Decorruptor-CM: Accelerate DPM to CM}
\label{Section:CM}
Motivated by consistency distillation introduced in Sec.~\ref{sec:cm}, we train a distilled variant for faster inference. The visual representations of our Decorruptor-CM model are illustrated in Fig. \ref{fig:proposed_pipeline} (b). Following the latent consistency distillation training protocol of LCM~\cite{luo2023latent}, we train Decorruptor-CM by minimizing the objective:
\begin{equation}
    \mathcal{L}_{LCD} (\theta, \theta^-; \Psi) = \mathbb{E}_{z_t, \omega, n} \left[ d \left( f_{\theta}(z_{t_{n+1}}, \omega, c, t_{n+1}), f_{\theta^-}(\hat{z}_{t_n}^{\Psi, \omega_I, \omega_{T}}, \omega, c, t_n) \right) \right],  
    \label{eq:LCD}
\end{equation}
where $f_{\theta}$ represents the student consistency model, $f_{\theta^-}$ denotes EMA of $f_{\theta}$, $\omega$ contains two guidance scales $\omega_T$ and $\omega_I$, and $c$ contains two conditions $c_T$ and $c_I$. EMA model's input $\hat{z}_{t_n}^{\Psi, \omega_I, \omega_{T}}$ is a prediction of the Decorruptor-DPM using PF-ODE solver augmented with multi-modal guidances:
\begin{equation}
    \begin{aligned}
        \hat{z}_{t_n}^{\Psi, \omega_I, \omega_{T}} - z_{t_{n+1}} & \approx \Psi(z_{t_{n+1}}, t_{n+1}, t_n, \emptyset, \emptyset) \\
        & + \omega_{I} (\Psi(z_{t_{n+1}}, t_{n+1}, t_n, c_I, \emptyset) - \Psi(z_{t_{n+1}}, t_{n+1}, t_n, \emptyset, \emptyset)) \\
        & + \omega_T (\Psi(z_{t_{n+1}}, t_{n+1}, t_n, c_{I}, c_{T}) - \Psi(z_{t_{n+1}}, t_{n+1}, t_n, c_{I}, \emptyset)).
        \label{eq_lcm_ip2p}
    \end{aligned}
\end{equation}
We train on the same dataset employed during the DPM training phase. While previous work~\cite{luo2023latent} has only augmented PF-ODE with text guidance, we introduce augmentation with multi-modal guidance as described in Eq.~(\ref{eq_lcm_ip2p}). 

In integrating the CFG scales $\omega_T$ and $\omega_I$ into the LCM, we employ Fourier embedding for both scales, following conditioning mechanisms of previous works~\cite{luo2023latent, meng2023distillation}. We use the zero-parameter initialization \cite{zhang2023adding} for stable training. We sample the embeddings $w_I$ and $w_T$, with a dimension of $768$, and multiply each other for U-Net conditioning. This approach allows the two variables to act as independent conditions during training, facilitating the multi-modal conditionings. Following Luo~\etal~\cite{luo2023latent}, $\omega_T^\text{min}$ and $\omega_T^{\text{max}}$ are set to $5.0$ and $15.0$, and we set $\omega_I^{\text{min}}$ and $\omega_I^{\text{max}}$ to be $1.0$ and $1.5$, respectively. It is worth noting that, unlike DPM, the integration of a learnable guidance scale obviates the need for image guidance scale scheduling. 

\subsection{Overall Process}
\label{Section:Overall}
The overall process for our efficient input updating TTA is as follows. First, complete the training of Decorruptor before TTA. 
Next, upon receiving the input image $x_0$, obtain the edited image $\hat{x}_0$ through Decorruptor. 
Finally, following the protocol of Gao~\etal~\cite{dda}, perform an ensemble to obtain the final prediction to capitalize on the classifier's knowledge from the target domain. 
It simply averages two predictions of $x_0$ and $\hat{x}_0$:
\begin{equation}
y^{pred} = 0.5 * (p_{\phi}(y|x_0) + p_{\phi}(y|\hat{x}_0)),
\label{eq:ensemble}
\end{equation}
where $y^{pred}$ means the final prediction of our TTA method and $p_{\phi}(x)$ means the probabilistic prediction of input $x$ by the pre-trained classifier $p_{\phi}$. In contrast to~\cite{dda}, we utilize probabilistic output after the softmax layer rather than logits for ensembling.

\section{Experimental Results}
In this section, we validate Decorruptor quantitatively and qualitatively, comparing it with the baseline and demonstrating its extensibility to other tasks.
\subsection{Setup}
\subsubsection{Benchmarks}
ImageNet-C~\cite{imagenet-c} is a benchmark with 15 types of algorithmically generated corruptions in four categories: noise, blur, weather, and digital, applied to the ImageNet~\cite{deng2009imagenet} dataset. ImageNet-$\bar{\mathrm{C}}$\cite{imagenet-c-bar} includes 10 perceptually different corruptions. To assess our method’s effectiveness in the video domain, we use UCF101-C\cite{lin2023video}, a benchmark with corruptions applied to UCF101~\cite{soomro2012ucf101}, which has a different distribution from ImageNet.
\subsubsection{Baselines}
For comparison, we utilized three baselines unaffected by compositions (e.g., label shifts, batch size). DiffPure~\cite{nie2022diffusion} employs diffusion for adversarial defense. DDA~\cite{dda} uses noise injection and denoising with an in-domain pre-trained DDPM, inspired by ILVR~\cite{choi2021ilvr}, and prevents catastrophic forgetting through self-ensembling. MEMO~\cite{zhang2022memo} updates the model using TTA with multiple data augmentations on a single input.

\subsubsection{Architectures}
We conducted the evaluation using ResNet50~\cite{he2016deep}, the most standard and lightweight network. Subsequently, to assess consistent performance improvement across various architectures, we followed the protocol of the Gao \etal~\cite{dda}: evaluating methods using the advanced forms of transformer structures (Swin-T, B~\cite{liu2021swin}) and convolution networks (ConvNeXt-T, B~\cite{liu2022convnet}).

\subsubsection{Data Preparation and Model Training}
For corruption crafting, we use the dataset provided by PIXMIX of fractals~\cite{nakashima2022can} and feature visualizations~\cite{baradad2021learning}. In each mixing operation in PIXMIX, we further apply SimSiam~\cite{chen2021exploring} transform, and various mixing sets. For model training, we initialize our model as a Stable Diffusion v$1.5$ model, and we follow the settings of IP2P for instruction fine-tuning considering image conditioning. We use ImageNet training data, with a size image of $256\times256$. We use a total batch size of $192$ for training the model for $30,000$ steps. This training takes about $2$ days on $8$ NVIDIA A40 GPUs, and we set the universal instruction as `\texttt{Clean the image}' for every clean-corruption pair while training. After training the DPM model, to accelerate DPM to CM, we further conduct distillation training for $24$ A40 GPU hours. Empirically, we found the distillation training for CM converges faster than training DPM. 
\begin{figure*}[t!]
    \centering
    \includegraphics[width=\linewidth]{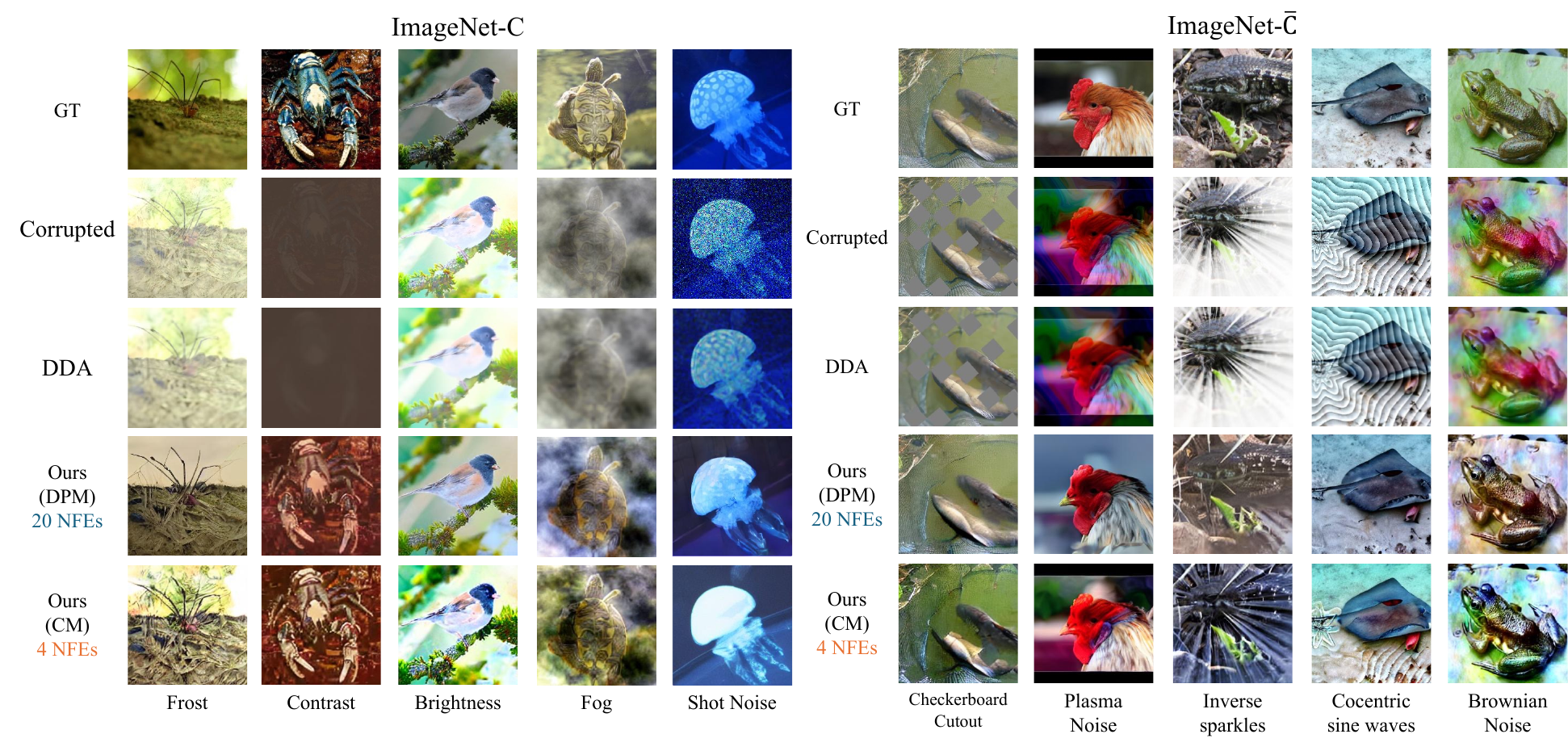} 
    \caption{Illustration of the results of corruption editing for various corruptions at severity $5$. Consequently, we have verified that our Decorruptor-DPM and CM generally enable effective editing for test-time corruptions.}
    \label{fig:comparative_result}
\end{figure*}
\subsection{Quantitative Evaluation}
In Table~\ref{table:performance_comparison}, we compare Decorruptor and other methods for ResNet-50. The memory of DDA and Decorruptor is the sum of the diffusion model and classifier memory. 4$\times$Decorruptor-CM represents marginalizing predictions of 4 edited samples using 4-step inference. Decorruptor shows the fastest runtime, highest performance, and least GPU memory consumption. Decorruptor-CM reduces runtime by over 100 times compared to DDA and is even faster than MEMO, a TTA method without diffusion. Decorruptor surpasses other baselines, especially on ImageNet-$\bar{\mathrm{C}}$, improving performance by 17.7$\%$ compared to DDA.

Tables~\ref{table:imagenet_c_accuracy} and~\ref{table:comparative_results} show Decorruptor’s superior performance across various classifiers and datasets. In ImageNet-C, 8$\times$Decorruptor-CM consistently outperforms DDA with shorter runtime than Decorruptor-DPM. DiffPure and DDA perform worse than the source-only model on ImageNet-$\bar{\mathrm{C}}$ with Swin-T and ConvNeXt-T. This suggests that CM enables multiple ensembles quickly, significantly improving performance. A detailed ensemble analysis is in Appendix B.7.
\begin{table}[t!]
\centering
\setlength{\tabcolsep}{4pt}
\renewcommand{\arraystretch}{1.1}
\caption{Comparisons with baselines on Imagenet-C dataset with ResNet-50. The bold and underlined values represent the best and second-best results, respectively. All performance metrics were measured using a single L40 GPU.}
\resizebox{\textwidth}{!}{
\begin{tabular}{lcccc}
\hline
Method          & Runtime (s/sample)\textbf{$\downarrow$} & Memory (MB)\textbf{$\downarrow$} & IN-C Acc. (\%)\textbf{$\uparrow$} & IN-$\bar{\mathrm{C}}$ Acc. (\%)\textbf{$\uparrow$} \\
\hline
MEMO     & \underline{0.41}      & 7456   & 24.7 & -       \\
DDA & 19.5      & 10320~+~2340   & 29.7 & 29.4       \\
\rowcolor{cyan!10}
Decorruptor-DPM      & 0.42      & \textbf{4602~+~2340}   & \underline{30.5} & \underline{41.8}       \\
\rowcolor{cyan!10}
4$\times$Decorruptor-CM      & \textcolor{black}{\textbf{0.14}}      & \underline{4958~+~2383}  & \textbf{32.8} & \textbf{47.1} \\
\hline
\end{tabular}
}
\label{table:performance_comparison}
\end{table}
\begin{table}[t!]
\centering
\setlength{\tabcolsep}{4pt}
\renewcommand{\arraystretch}{1.1}
\caption{Comparisons with baselines on ImageNet-C at severity level 5 in terms of
the average accuracy of 15 corruptions (\%). The bold and underlined values represent the best and second-best results, respectively.}
\resizebox{1.0\textwidth}{!}{
\begin{tabular}{lccccc}
\hline
Method          & ResNet-50 & Swin-T & ConvNeXt-T & Swin-B & ConvNeXt-B  \\
\hline
Source-Only     & 18.7      & 33.1   & 39.3 & 40.5 & 45.6       \\
MEMO \tiny{\textcolor{black}{(0.41s)}} & 24.7 & 29.5   & 37.8 & 37.0 & 45.8       \\
DiffPure \tiny{\textcolor{black}{(27.3s)}}      & 16.8     & 24.8   & 28.8  & 28.9  & 32.7     \\
DDA \tiny{\textcolor{black}{(19.5s)}}     & 29.7     & \underline{40.0}   & \underline{44.2} & {44.5}  & \underline{49.4}     \\
\rowcolor{cyan!10}
Decorruptor-DPM \tiny{\textcolor{blue}{(0.42s)}}     & {30.5}      & 37.8   & 42.2  & 42.5  & 46.6 \\
\rowcolor{cyan!10}
4$\times$Decorruptor-CM \tiny{\textcolor{blue}{(0.14s)}}      & \underline{32.8}       & {39.7}  & {44.0}  & \underline{44.7} & {48.6} \\
\rowcolor{cyan!10}
8$\times$Decorruptor-CM \tiny{\textcolor{blue}{(0.25s)}}      & \textbf{34.2}       & \textbf{41.1}  & \textbf{45.2}  & \textbf{46.1} & \textbf{49.8} \\
\hline
\end{tabular}
}
\label{table:imagenet_c_accuracy}
\end{table}
\begin{table}[h!]
\centering
\setlength{\tabcolsep}{4pt}
\renewcommand{\arraystretch}{1.1}
\caption{Comparisons with baselines on ImageNet-$\bar{\mathrm{C}}$ at severity level 5 in terms of the average accuracy of 15 corruptions (\%). The bold and underlined values represent the best and second-best results, respectively.}
\resizebox{0.65\textwidth}{!}{\begin{tabular}{lccc}
\hline
Method          & ResNet-50 & Swin-T & ConvNeXt-T \\
\hline
\rowcolor{gray!10}
Source-Only     & 25.8      & {44.2}   & {47.2}       \\
\hline
DiffPure & 19.8 \tiny{\textcolor{red}{(-6.0)}}    & 28.5 \tiny{\textcolor{red}{(-15.7)}}   & 32.1 \tiny{\textcolor{red}{(-15.1)}}       \\
DDA      & {29.4} \tiny{\textcolor{blue}{(+3.6)}}     & 43.8 \tiny{\textcolor{red}{(-0.4)}}  & 46.3 \tiny{\textcolor{red}{(-0.9)}}      \\
\rowcolor{cyan!10}
Decorruptor-DPM      & \underline{41.8} \tiny{\textcolor{blue}{(+16.0)}}     & \underline{52.5} \tiny{\textcolor{blue}{(+8.3)}} & \underline{55.0} \tiny{\textcolor{blue}{(+7.8)}}\\
\rowcolor{cyan!10}
4$\times$Decorruptor-CM      & \textbf{47.1} \tiny{\textcolor{blue}{(+21.3)}}     & \textbf{55.8} \tiny{\textcolor{blue}{(+11.6)}} & \textbf{58.6} \tiny{\textcolor{blue}{(+11.4)}}\\
\hline
\end{tabular}}
\label{table:comparative_results}
\end{table}


\subsection{Analysis on Decorruptor}
\label{sec:analysis}
\subsubsection{Comparisons with DDA}
We employ the LPIPS~\cite{zhang2018unreasonable} metric to measure the image-level perceptual similarity. As seen in Table~\ref{tab:lpips}, Decorruptor-DPM shows lower similarity with ImageNet-C than DDA, but closer similarity with ImageNet. This suggests that Decorruptor performs more edits on corrupted images than DDA, and the edited images become cleaner. Moreover, in Fig. \ref{fig:comparative_result}, we showcase the qualitative results of Decorruptors and DDA for a range of corruptions on ImageNet-C and ImageNet-$\bar{\mathrm{C}}$. As a result, our Decorruptor consistently outperforms DDA for all of the corruption editing. 

\subsubsection{Orthogonality with Model Updating Methods}
We conducted experiments to explore the feasibility of combining Decorruptor with model updating methods. We compared the performance by ensembling predictions of images edited with Decorruptor-DPM to the existing model updating method. As seen in Table~\ref{tab:orthogonality}, Decorruptor demonstrated performance improvements for both TENT~\cite{tent} and DeYO~\cite{deyo}. These outcomes suggest the potential for an advanced TTA approach through the integration of model updating and input updating methods. Further results are represented in the Appendix B.2. 

\subsubsection{Performance Trade-Off Analysis of Deccorruptor-CM}
Table~\ref{tab:ensembling} reports the experimental results for various choices of Decorruptor-CM. When editing the input image only once for prediction, it is fast but the performance decreases by about 2-3\% compared to Decorruptor-DPM. However, obtaining four edited images for each input and using their average prediction in Eq.~(\ref{eq:ensemble}) leads to higher accuracy than DPM, regardless of the architecture. 
Notably, even for the same corrupted image, Decorruptor allows diverse edits towards different clean images.
The increasing performance gap between 1 step and 4 steps as the model size grows suggests that the quality of images is inferred to be better with 4 steps. 

\begin{figure}[t!]
\centering

\begin{minipage}[t]{0.3\textwidth}
  \centering
  \setlength{\tabcolsep}{4pt}
  \captionof{table}{LPIPS scores with clean and corrupted images.}
  \resizebox{1.0\textwidth}{!}{
    \begin{tabular}{lcc}
      \toprule
      LPIPS & IN-C($\uparrow$) & IN($\downarrow$) \\
      \midrule
      DDA & 0.421 & 0.608\\
      \rowcolor{cyan!10}
      Decorruptor-DPM & \textbf{0.575} & \textbf{0.573} \\
      \bottomrule
    \end{tabular}
  }
  \label{tab:lpips}
\end{minipage}%
\hfill
\begin{minipage}[t]{0.3\textwidth}
  \centering
  \captionof{table}{Orthogonality with model updates.}
  \resizebox{1.0\textwidth}{!}{
    \begin{tabular}{lc}
      \toprule
      & Avg. Acc (\%) \\
      \midrule
      TENT & 43.02 \\
      \rowcolor{cyan!10}
      + Decorruptor-DPM & \textbf{45.52} \\
      \midrule
      DeYO & 48.61 \\
      \rowcolor{cyan!10}
      + Decorruptor-DPM & \textbf{49.50} \\
      \bottomrule
    \end{tabular}
  }
  \label{tab:orthogonality}
\end{minipage}%
\hfill
\begin{minipage}[t]{0.32\textwidth}
  \centering
  \setlength{\tabcolsep}{4pt}
  \captionof{table}{Variants of Decorruptor-CM.}
  \resizebox{1.0\textwidth}{!}{
    \begin{tabular}{lcc}
      \toprule
      Decorruptor & ResNet-50 & Swin-T \\
      \midrule
      \rowcolor{gray!10}
      DPM \textcolor{blue}{\tiny{(0.42s)}} & 30.5 & 37.8 \\
      \midrule
      CM (1step) \textcolor{blue}{\tiny{(0.05s)}} & 26.8 & 34.5 \\
      4$\times$CM (1step) \textcolor{blue}{\tiny{(0.08s)}} & 32.6 & 38.7 \\
      CM (4step) \textcolor{blue}{\tiny{(0.10s)}} & 27.5 & 35.6 \\
      4$\times$CM (4step) \textcolor{blue}{\tiny{(0.14s)}} & \textbf{32.8} & \textbf{39.7} \\
      \bottomrule
    \end{tabular}
  }
  \label{tab:ensembling}
\end{minipage}
\end{figure}

\begin{table}[t!]
\centering
\caption{Performance comparisons based on different source model in OOD datasets.}
\resizebox{1.0\textwidth}{!}{
    \begin{tabular}{
    l|c
    c
    >{\columncolor{cyan!10}}c 
    >{\columncolor{cyan!10}}c 
    |
    c
    c
    >{\columncolor{cyan!10}}c  
    >{\columncolor{cyan!10}}c
    }
    \toprule
    & Source & + DDA & + DPM & + 4$\times$CM & PIXMIX & + DDA & + DPM & + 4$\times$CM \\
    \midrule
    VISDA-2021 acc (\%) & 35.7 & 40.2 & \underline{40.9} & \textbf{42.0} & 44.0 & 45.4 & \underline{45.6} & \textbf{46.1}  \\
    ImageNet-A acc (\%)        & 0.0  & 0.5  & \underline{1.9} & \textbf{2.7}  & 6.3 & 5.2 \scriptsize{\textcolor{red}{(-1.1)}} & \underline{8.1}  & \textbf{9.8}  \\
    \bottomrule
    \end{tabular}
}
\label{tab:ood_performance}
\end{table}%
\begin{table}[t!]
\centering
\setlength{\tabcolsep}{4pt}
\renewcommand{\arraystretch}{1.1}
\caption{Performance comparisons based on different types of data augmentation methods used in the corruption modeling scheme.}
\resizebox{0.75\textwidth}{!}{
\begin{tabular}{lcc}
\hline
Method          & IN-C Acc (\%)\textbf{$\uparrow$} & VISDA-2021 Acc (\%)\textbf{$\uparrow$} \\
\hline
PIXMIX & 28.2      & 38.2 \\
SimSiam      & 20.4      & 37.0 \\
\rowcolor{cyan!10}
PIXMIX + SimSiam (Ours)     & 30.5      & 40.9 \\
\hline
\end{tabular}
}
\label{table:aug_ab}
\end{table}
\subsubsection{Analysis of OOD Generalization Performance}
To demonstrate Decorruptor’s OOD generalization capabilities, we used the VISDA-2021 dataset~\cite{visda_2021}, which includes ImageNet-C, ImageNet-R, ObjectNet, and ImageNet-A~\cite{imagenet_a}. As shown in Table~\ref{tab:ood_performance}, our method outperforms both the source model and DDA on every benchmark dataset. Additionally, using Decorruptor with a single robust source model (PIXMIX) results in further performance gains, while DDA shows a performance drop on ImageNet-A. The reasons for our performance improvements in OOD are: 1) Initialization: Initializing Decorruptor with Stable Diffusion, pre-trained on the 5-billion-scale LAION dataset, enhances generalization on OOD datasets, even after fine-tuning on ImageNet. 2) Corruption Modeling Scheme: This scheme improves robustness against unseen corruption by expanding the model’s manifold by recovering corrupted images to clean images. Table~\ref{table:aug_ab} shows a noticeable performance difference when using PIXMIX and SimSiam, with the combination maximizing performance. This indicates that our corruption modeling scheme with effective data augmentations is beneficial for OOD generalization.


\subsubsection{Multi-Modal Guidance Conditioning Analysis}
We describe the benefits of using a learnable image guidance scale in CM. LCM~\cite{luo2023latent} demonstrates fast convergence and significant performance improvements in distillation by using learnable guidance scales on SD, focusing on a learnable text guidance \( w_T \). However, we found that directly implementing LCM results in undesirable outcomes for Decorruptor, which receives both text and image inputs. Recognizing the importance of conditioning the image guidance scale \( w_I \), we proposed a new learnable multi-modal guidance \( w = w_I \cdot w_T \). Detailed results are presented in Appendix C.1.
\subsection{Video Test-Time Adaptation}
\label{subsec:videotta}
Decorruptor demonstrates significantly improved runtime compared to DDA, making it practical for both image and video domains. We evaluated Decorruptor on a corrupted video dataset~\cite{lin2023video}, applying Text2Video-Zero~\cite{khachatryan2023text2video} for temporally consistent frames. Text2Video-Zero uses cross-frame attention from the first frame across the sequence for coherent editing. Detailed quantitative results are in Appendix B.4.

\section{Conclusions}
The existing diffusion-based image-level updating TTA approach is robust to data order and batch size variations but is impractical for real-world usage due to its slow processing speed. In response, we propose Decorruptor-DPM, leveraging a latent diffusion model for efficient memory and time utilization. Through fine-tuning via our novel corruption modeling scheme, Decorruptor-DPM possesses the capability to edit corrupted images. Additionally, we introduce Decorruptor-CM, employing consistency distillation to accelerate input updates further. Decorruptor surpasses the baseline diffusion-based approach in speed by 100 times while delivering superior performance. 
\clearpage
\newpage
\subsubsection{Acknowledgement}
This work was partly supported by Institute of Information \& communications Technology Planning \& Evaluation (IITP) grant funded by the Korea government(MSIT) [NO.RS-2021-II211343, Artificial Intelligence Graduate School Program (Seoul National University)], the National Research Foundation of Korea (NRF) grant funded by the Korea government (MSIT) (No.2022R1A3B1077720) and the BK21 FOUR program of the Education and the Research Program for Future ICT Pioneers, Seoul National University in 2024.

%
%
\bibliographystyle{splncs04}
\bibliography{main}
\appendix

\section{Further Explanations}
\subsection{TTA with Data Augmentation Approaches}
Data augmentation is being used across various fields to enhance robustness against distribution shifts, including supervised~\cite{zhang2017mixup, yun2019cutmix}, semi-supervised~\cite{sohn2020fixmatch}, and self-supervised learning~\cite{chen2021exploring}. Similarly, in TTA, MEMO~\cite{zhang2022memo} applies multiple data augmentations to a single input. Notably, it utilizes $64$ data augmentations on test-time input to minimize marginal entropy and enables more stable adaptation than TENT~\cite{tent}. Here, MEMO applies data augmentation to the input and tunes the classifier by minimizing the averaged prediction over augmentations. In contrast, Decorruptor uses augmentation when training the diffusion model for robustness against distribution shifts before adaptation.

\subsection{Detailed Contributions of Decorruptor}
We clarify our Decorruptor is a classifier-agnostic generator that modifies corrupted images into clean images. 
Decorruptor allows for obtaining stable performance through the ensemble of multiple decorrupted images. Moreover, it can also remove corruption from images with out-of-distribution (OOD) classes, making them usable for downstream tasks. 
This is supported by VideoTTA results (Section 5.4) in the main text. 
These are key differences from other EM-based TTA approaches (\eg, EATA, SAR, and DeYO) and a single robust model (\eg, PIXMIX). Following Eq. (10), the final prediction is obtained by ensembling the predictions of the generated clean images with the original prediction. This indicates that Decorruptor can be applied \textit{orthogonally} with other TTA methods that modify the original prediction. Note that the single Decorruptor checkpoint was utilized \textit{across all datasets}, methods (\eg, EATA and PIXMIX), and tasks (\eg, image and video classification), demonstrating its versatility. Furthermore, Decorruptor achieves a threefold increase in speed and superior performance compared to the data augmentation-based model updating baseline. 

\subsection{Justification/Implication of Universal Prompt:}
We chose the general text prompt (\texttt{Clean the Image}) to handle \textit{any} unknown distribution shifts at test time (\ie, corruption levels and types), 
Other valid prompts (\eg, \texttt{Decorrupt the image}) will also work while they are fixed during training and inference time. To clarify it, since this prompt is fixed, our Decorruptor can be considered an image-to-image translation model that reverts corrupted images to their clean image counterparts.

\subsection{Pseudo-codes}
We provide a pseudo-code in Algorithm \ref{algorithm2} for training Decorruptor-CM with consistency distillation~\cite{song2023consistency}. Note, following LCM~\cite{luo2023latent}, two distinct timesteps $t_n$ and $t_{n+k}$, which are $k$ steps apart, are randomly selected and the same Gaussian noise $\epsilon$ are applied. The generated noisy latents $z_{t_n}$ and $z_{t_{n+k}}$ are represented as follows:
\begin{equation*}
    z_{t_{n+k}} = \alpha(t_{n+k})z + \sigma(t_{n+k})\epsilon, \quad z_{t_n} = \alpha(t_n)z + \sigma(t_n)\epsilon.
\end{equation*}
Note, following our multi-modal guidance scheme described in the main text, the self-consistency property can be held during distillation and the skipping-step technique can also be used. In the following, we append the inference pseudo-code of both Decorruptor-DPM and CM as described in Algorithm \ref{algorithm1}.
\begin{algorithm}
\caption{Decorruptor-CM Training}
\begin{algorithmic}[1]
\State \textbf{Input:} Given dataset $\mathcal{D}^{(p)}$, distance metric $d(\cdot, \cdot)$, pre-trained model parameter $\theta$,learning rate $\eta$, EMA coeff $\mu$, noise schedule $\alpha(t)$, $\sigma(t)$, multi-modal guidance scale: $[w_{I,\min}, w_{I,\max}]$ and $[w_{T,\min}, w_{T,\max}]$, skipping interval $k$, and encoder $E(\cdot)$
\State Encode paired clean/corrupt data into the latent space: $\mathcal{D}^{(p)}_z = \{(z_{c1}, z_{co}, c) \mid z = E(x), (x_{cl}, x_{co}, c) \in \mathcal{D}^{(p)}\}$
\State $\theta^- \gets \theta$ $\hfill \triangleright$ Initialization
\Repeat
    \State Sample $(z_{cl}, z_{co}, c) \sim \mathcal{D}^{(p)}_z$, $n \sim \mathcal{U}[1, N - k]$
    \State Sample $\epsilon \sim \mathcal{N}(0, I)$, $w_I$ and $w_T$
    \State $z_{t_{n+k}} \gets \alpha(t_{n+k})z + \sigma(t_{n+k})\epsilon$ 
    \State $z_{t_n} \gets \alpha(t_n)z + \sigma(t_n)\epsilon$
    \State Minimize Eq. (4)
    \State $\theta \gets \theta - \eta \nabla_\theta \mathcal{L}(\theta, \theta^-)$
    \State $\theta^- \gets \text{stopgrad}(\mu \theta^- + (1 - \mu) \theta)$
\Until{convergence}
\end{algorithmic}
\label{algorithm2}
\end{algorithm}

\begin{algorithm}
    \caption{Decorruptor-DPM/CM Inference}
    \begin{algorithmic}[1]
    \State \textbf{Input:} Text and image guidance scales $\omega_{T}$ and $\omega_{I}$, Given text $c_{T}$ and corrupted image $c_{I}$, Noise schedule $\alpha(t)$, $\sigma(t)$, Decoder $D(\cdot)$ 
    \State $ts$: Diffusion timesteps (20 for DPM, 4 for CM), $T$: maximum  timesteps (1000) 
    \State $\epsilon_{\theta}$: Pre-trained DPM or CM 
    \State Sample $\hat{z}_T \sim \mathcal{N}(0;I)$,
    $z \gets \epsilon_\theta(\hat{z}_T, c_{T}, c_{I}, T)$ 
    \For{$t \gets ts \ldots 1$}\hfill $\triangleright$ sequence of timesteps
        \State $\hat{z}_{t} \sim \mathcal{N}(\alpha(t) z; \sigma^2(t) I)$
        \State Eq (7), (9) for DPM, CM $\hfill \triangleright$ multi-modal guidance
        \State $z \gets \epsilon_\theta(\hat{z}_t, c_{T}, c_{I}, t)$ 
    \EndFor
    \State $\hat{x}_0 \gets D(z)$ \hfill $\triangleright$ decoding latent to decorrupted image
    \State \textbf{return} $\hat{x}_0$
    \end{algorithmic}
    \label{algorithm1}
\end{algorithm}

\section{Additional Results}
\subsection{Detailed Results of Image Corruption Editing}
\begin{figure*}[h!]
\includegraphics[width=\linewidth]{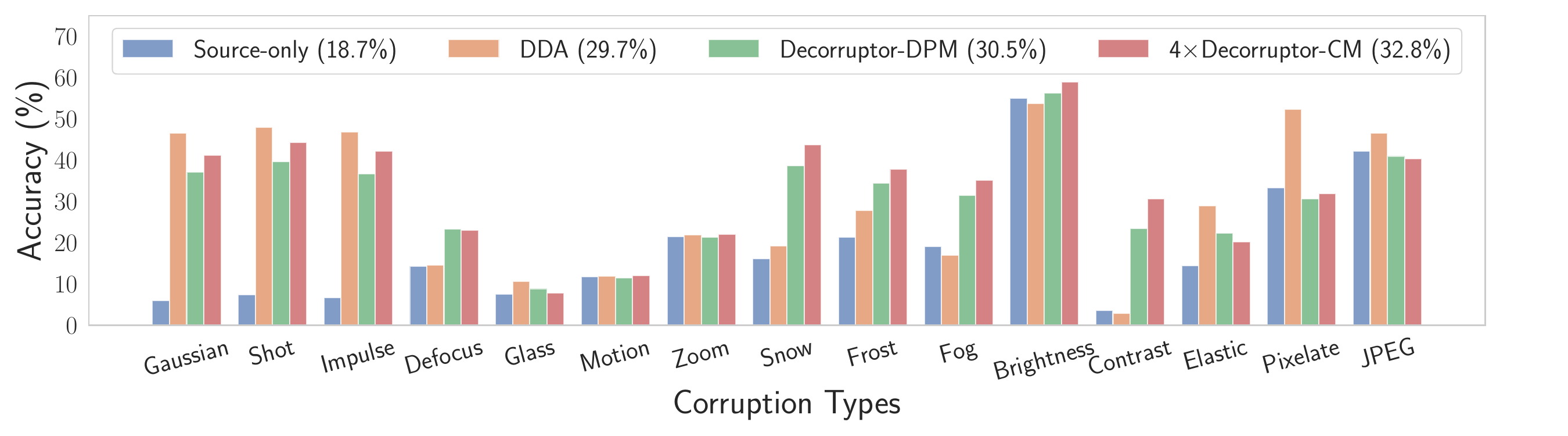} 
    \caption{Bar graph of comparisons for performances between DDA and our Decorruptors on ImageNet-C using ResNet50.}
    \label{fig:comparative_bar}
\end{figure*}

Tables~\ref{tab:imagenet-c-detail} and~\ref{tab:imagenet-c-bar-detail} present detailed performance results of Decorruptor on ImageNet-C~\cite{imagenet-c} and ImageNet-$\bar{\mathrm{C}}$~\cite{imagenet-c-bar}, respectively, using ResNet-50~\cite{he2016deep} as the architecture. As shown in Table~\ref{tab:imagenet-c-detail}, Decorruptor demonstrates significant performance improvement for Noise and Weather corruptions. Moreover, in Fig.~\ref{fig:comparative_bar}, 4$\times$Decorruptor-CM shows performance improvements over the source-only case in all corruptions except for pixelate and jpeg, and it outperforms DPM in most corruptions. 

DDA~\cite{dda} also shares the limitation of not being able to properly edit some corruptions. Addressing this issue is crucial for the effectiveness of the input updating TTA~\cite{tent} method. As illustrated in Table~\ref{tab:imagenet-c-bar-detail}, Our Decorruptor shows consistent improvement for all corruptions in ImageNet-$\bar{\mathrm{C}}$. Commonly, ensembling more edited images always presents performance improvement for all corruptions in ImageNet-C and ImageNet-$\bar{\mathrm{C}}$. 


\begin{table*}[h]
    
    \setlength{\tabcolsep}{4pt}
    \caption{Detailed results on ImageNet-C at severity level 5 regarding accuracy (\%). The \textbf{bold} value signifies the top-performing result.}
\newcommand{\tabincell}[2]{\begin{tabular}{@{}#1@{}}#2\end{tabular}}
 \begin{center}
  \begin{threeparttable}
 \LARGE
 
    \resizebox{1.0\linewidth}{!}{
 	\begin{tabular}{l|ccc|cccc|cccc|cccc|c}
 	\multicolumn{1}{c}{} & \multicolumn{3}{c}{Noise} & \multicolumn{4}{c}{Blur} & \multicolumn{4}{c}{Weather} & \multicolumn{4}{c}{Digital}  \\
 	 ImageNet-C & Gauss. & Shot & Impul. & Defoc. & Glass & Motion & Zoom & Snow & Frost & Fog & Brit. & Contr. & Elastic & Pixel & JPEG & Avg.  \\
    \cmidrule{1-17}
        ResNet-50 (Source-only) & 6.1                        & 7.5                      & 6.7                        & 14.4                       & 7.6                       & 11.8                       & 21.4                     & 16.2                     & 21.4                      & 19.1                    & 55.1                      & 3.6                        & 14.5                        & \textbf{33.3}                      & \textbf{42.1}                     & 18.7  \\ 
        ~~$\bullet~$Decorruptor-DPM & 37.1                       & 39.7                     & 36.7                       & 23.3                       & \textbf{8.9}                      & 11.5                       & 21.4                     & 38.7                     & 34.4                      & 31.5                    & 56.3                      & 23.5                       & \textbf{22.4}                        & 30.6                      & 41.0                     & 30.5  \\
        ~~$\bullet~$Decorruptor-CM & 30.3                       & 33.4                     & 30.7                       & 19.5                       & 7.4                      & 11.6                       & 20.8                     & 33.2                     & 29.8                     & 28.5                    & 56.0                      & 22.1                       & 17.9                        & 30.6                      & 40.2                     & 27.5  \\
        ~~$\bullet~$4$\times$Decorruptor-CM & 41.2                       & 44.3                     & 42.2                       & 23.1                       & 7.8                      & 12.1                       & 22.0                     & 43.7                     & 37.8                      & 35.2                    & 58.9                      & 30.6                       & 20.3                        & 31.9                      & 40.4                     & 32.8  \\ 
        \rowcolor{cyan!10}~~$\bullet~$8$\times$Decorruptor-CM & \textbf{44.1}                       & \textbf{46.7}                     & \textbf{44.9}                       & \textbf{24.0}                       & 8.0                      & \textbf{12.4}                       & \textbf{22.5}                     & \textbf{46.2}                     & \textbf{40.3}                      & \textbf{36.4}                    & \textbf{59.7}                      & \textbf{32.6}                       & 20.6                        & 33.2                      & 41.1                     & \textbf{34.2} \\ 
	\end{tabular} 
        }
    \end{threeparttable}
    \end{center}
\vspace{-0.13in}
    \label{tab:imagenet-c-detail}
\end{table*}
\begin{table*}[h!]
    
    \setlength{\tabcolsep}{4pt}
    \caption{Detailed results on ImageNet-$\bar{\mathrm{C}}$ at severity level 5 regarding accuracy (\%). The \textbf{bold} value signifies the top-performing result.}
\newcommand{\tabincell}[2]{\begin{tabular}{@{}#1@{}}#2\end{tabular}}
 \begin{center}
  \begin{threeparttable}
 \LARGE
 
    \resizebox{1.0\linewidth}{!}{
 	\begin{tabular}{l|cccccccccc|c}
 	 ImageNet-$\bar{\mathrm{C}}$ & Blue. & Brown. & Caustic. & Checker. & Cocentric. & Inverse. & Perlin. & Plasma. & Single. & Sparkles & Avg.  \\
    \cmidrule{1-12}
        ResNet-50 (Source-only) & 23.7                        & 41.3                      & 37.7                        & 32.7                       & 4.2                       & 9.3                       & 46.3                     & 9.9                     & 4.6                      & 48.1                    & 25.8 \\ 
        ~~$\bullet~$Decorruptor-DPM & 38.6                       & 53.5                     & 45.3                       & 45.4                       & {31.5}                      & 26.6                       & 54.8                     & 34.0                     & 30.6                      & 58.1                    & 41.8 \\
        ~~$\bullet~$Decorruptor-CM & 37.4                       & 51.5                     & 43.5                       & 44.2                       & {27.1}                      & 21.3                       & 53.1                     & 29.3                     & 26.8                      & 56.7                    & 39.1 \\
        ~~$\bullet~$4$\times$Decorruptor-CM & 45.6                       & 57.4                     & 48.2                       & 51.5                       & {42.2}                      & 28.2                       & 57.5                     & 39.8                     & 39.6                      & 61.2                    & 47.1 \\
        \rowcolor{cyan!10}~~$\bullet~$8$\times$Decorruptor-CM & \textbf{47.2}                       & \textbf{58.2}                     & \textbf{49.2}                       & \textbf{53.4}                       & \textbf{43.2}                      & \textbf{29.4}                       & \textbf{58.6}                     & \textbf{41.2}                     & \textbf{40.4}                      & \textbf{62.2}                    & \textbf{48.3} \\
	\end{tabular} 
        }
    \end{threeparttable}
    \end{center}
\vspace{-0.13in}
    \label{tab:imagenet-c-bar-detail}
\end{table*}

\subsection{Comparisons with EM-Based TTA Methods}
In Table \ref{table:computation}, we present additional comparisons of our computational costs in terms of accuracy and memory compared to existing EM-based TTA methods. As a result, adding Decorruptor-4$\times$CM resulted in significant improvements across all approaches and datasets. In terms of efficiency, DDA incurs an additional runtime of 19.5s, whereas 4$\times$CM adds an additional runtime of 0.14s, which is only about three times the runtime of EM-based methods (0.05s). Decorruptor shows significant improvements compared to the state-of-the-art EM-based method DeYO (48.6\% $\rightarrow$ 51.6\%) and can be applied to other downstream tasks (\eg, VideoTTA), demonstrating its strong superiority.
\begin{table}[t]
\centering
\caption{Comparisons with TTA methods.}
\resizebox{1\textwidth}{!}{
    \begin{tabular}{
    l|c
    >{\columncolor{cyan!10}}c 
    >{\columncolor{cyan!10}}c 
    |
    c
    >{\columncolor{cyan!10}}c  
    >{\columncolor{cyan!10}}c 
    |
    c
    >{\columncolor{cyan!10}}c  
    >{\columncolor{cyan!10}}c 
    }
    \toprule
    & Source & + DPM & + 4$\times$CM & EATA & + DPM & + 4$\times$CM & SAR & + DPM & + 4$\times$CM  \\
    \midrule
    Runtime (s/sample) & 0.004 & + 0.42 & + 0.14 & 0.047 & + 0.42 & + 0.14 & 0.054 & + 0.42 & + 0.14 \\
    Memory (MB)        & 2,340  & + 4,602  & + 4,958 & 2,704  & + 4,602 & + 4,958 & 2,702  & + 4,602 & + 4,958 \\
    IN-C acc (\%)       & 18.0    & \underline{31.2}  & \textbf{33.8} & \underline{47.8}  & 47.5 & \textbf{51.6} & 44.0    & \underline{47.4} & \textbf{49.6} \\
    IN-$\bar{\mathrm{C}}$ acc (\%)       & 25.0    & \underline{41.6}  & \textbf{47.7} & 54.0  & \underline{57.6} & \textbf{59.7} & 49.9   & \underline{55.9} & \textbf{58.8} \\
    \bottomrule
    \end{tabular}
}
\label{table:computation}
\end{table}
\subsection{Quantitative Video Corruption Editing Results}
\begin{figure*}[t!]
    \centering
    \includegraphics[width=0.9\linewidth]{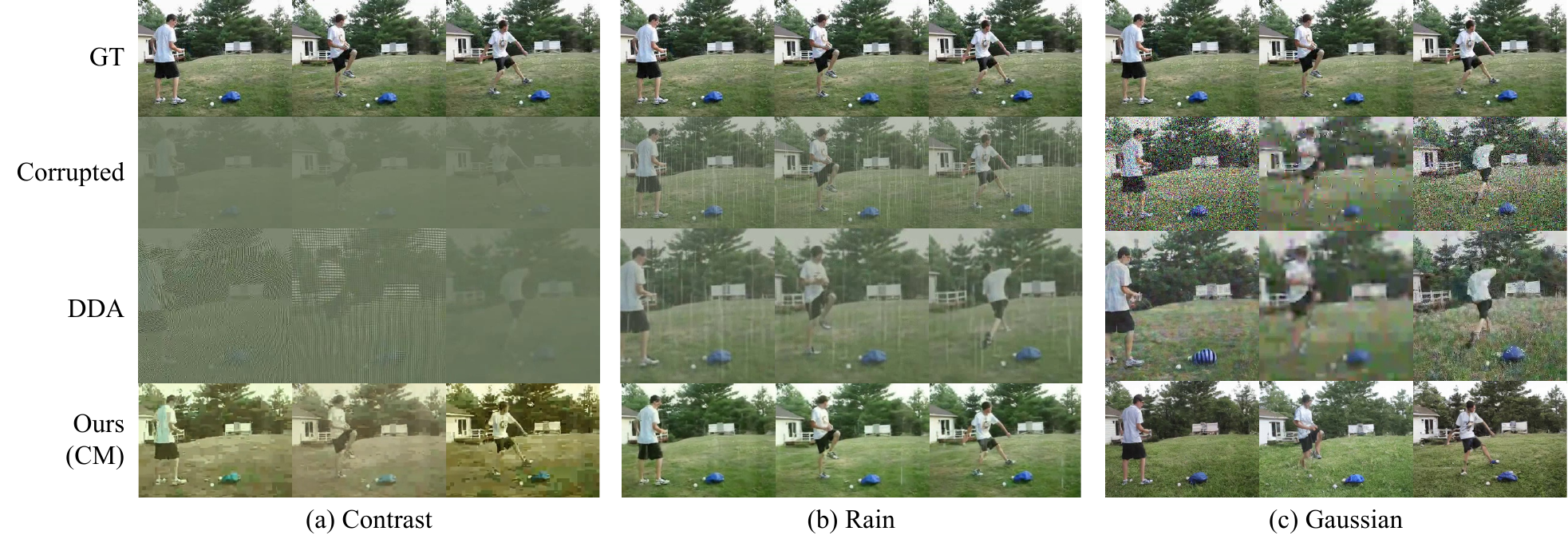} 
    \caption{Results of corruption editing for corrupted videos in UCF101-C.}
    \label{fig:video_editing}
\end{figure*}

In this section, we elaborate on the results obtained by applying Decorruptor-CM for video corruption editing. As shown in Fig.~\ref{fig:video_editing}, Decorruptor-CM outperformed DDA in corruption editing. For a 3-second input video, Decorruptor takes about 10 seconds, while DDA takes nearly 20 minutes. This demonstrates that Decorruptor is both highly effective and efficient for video corruption editing. For our experiments, we referred to the performance chart of ViTTA (see Table 2 in Lin~\etal~\cite{lin2023video}). The UCF-101C~\cite{lin2023video} dataset includes 3,783 corrupted videos for each type of corruption, covering a total of 12 different corruptions. The entire process of video decorruption was conducted using eight A40 GPUs and took about three days. The network used in the experiments was TANet~\cite{liu2021tam}. Instead of using an ensemble, we assessed the performance solely using the generated dataset when combining the model update method with our approach. The results are described in Table~\ref{tab:vidtta}. Ensembling with the source resulted in an average performance improvement of approximately 13\% compared to source-only inference. These findings suggest that our Decorruptor-CM can be effectively applied to video domains. Additionally, by applying our method with the TTA methodology, we observed an average performance improvement of about 3\% compared to ViTTA, particularly showing robust decorruption results against noise.
\begin{table}[h!]
\caption{Quantitative results for the UCF101-C dataset. Here, `Ours-Only' refers to results obtained from inference using only input updates. We further provide the results of combining our methodology with the baseline TTA method.}
\scalebox{0.7}{
\centering
\begin{tabular}{ccccccccccccccc}
\hline
Update & Methods & Gauss & Pepper & Salt & Shot & Zoom & Impulse & Defocus & Motion & JPEG & Contrast & Rain & H265.abr & Avg \\
\hline
& Source-Only & 17.92 & 23.66 & 7.85 & 72.48 & 76.04 & 17.16 & 37.51 & 54.51 & 83.40 & 62.68 & 81.44 & 81.58 & 51.35 \\
\rowcolor{cyan!10}
Data & Ours-Only & 42.43 & 54.24 & 33.01 & 85.83 & 75.83 & 56.25 & 37.82 & 58.33 & 85.77 & 74.83 & 85.85 & 81.97 & \textbf{64.34} \\
\hline
& NORM~\cite{schneider2020improving} & 45.23 & 42.43 & 27.91 & 86.25 & 84.43 & 46.31 & 54.32 & 64.19 & 89.19 & 75.26 & 90.43 & 83.27 & 65.77 \\
& DUA~\cite{mirza2022norm} & 36.61 & 33.97 & 22.39 & 80.25 & 77.13 & 36.72 & 44.89 & 55.67 & 85.12 & 30.58 & 82.66 & 78.14 & 55.34 \\
& TENT~\cite{tent} & 58.34 & 53.34 & 35.77 & 89.61 & \underline{87.68} & 59.08 & 64.92 & 75.59 & 90.99 & 82.53 & 92.12 & \textbf{85.09} & 72.92 \\
Model & SHOT~\cite{liang2020we} & 46.10 & 43.33 & 29.50 & 85.51 & 82.95 & 47.53 & 53.77 & 63.37 & 88.69 & 73.30 & 89.82 & 82.66 & 65.54 \\
& T3A~\cite{iwasawa2021test} & 19.35 & 26.57 & 8.83 & 77.19 & 79.38 & 18.64 & 40.68 & 58.61 & 86.12 & 67.22 & 84.00 & 83.45 & 54.17 \\
& ViTTA~\cite{lin2023video} & \underline{71.37} & \underline{64.55} & \underline{45.84} & \underline{91.44} & \textbf{87.68} & \underline{71.90} & \textbf{70.76} & \textbf{80.32} & \underline{}{91.70} & \textbf{86.78} & \textbf{93.07} & \underline{84.56} & 78.33 \\
\rowcolor{cyan!10}
All & ViTTA + Ours & \textbf{77.05} & \textbf{79.03} & \textbf{64.18} & \textbf{93.25} & 86.54 & \textbf{78.32} & \underline{65.72} & \underline{78.30} & \textbf{91.76} & \underline{86.41}  & \underline{92.25} & 83.58 & \textbf{81.37} \\
\hline
\end{tabular}}
\vspace{0.1in}
\label{tab:vidtta}
\end{table}
\subsection{Corruption Granularity}
\begin{figure*}[t!]
    \centering
    \includegraphics[width=0.8\linewidth]{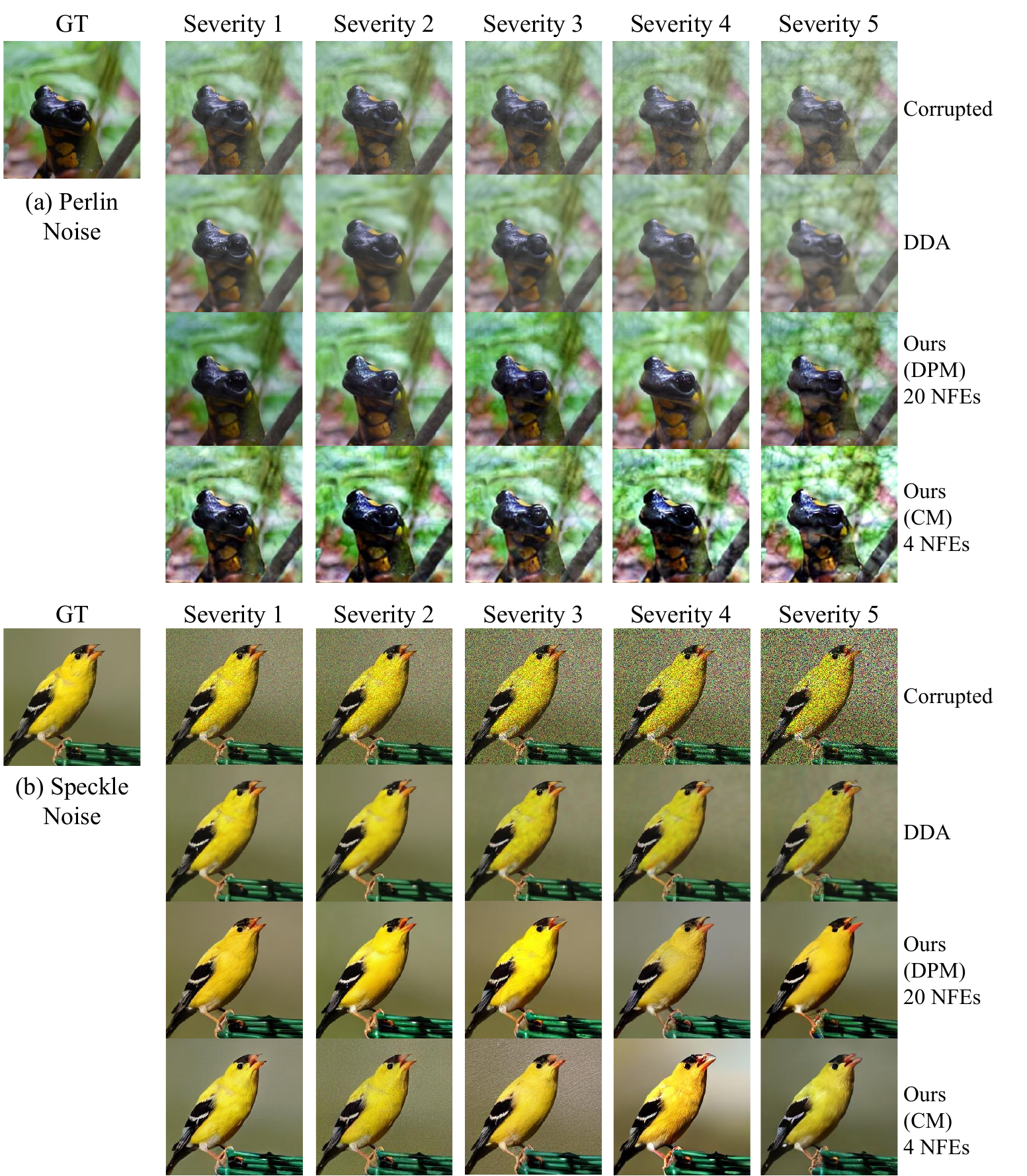} 
    \caption{Visualization of corruption editing results based on the granularity of severity for various corruptions.}
    \label{fig:severity_granulity}
\end{figure*}

Our proposed Decorruptor-DPM and CM methodologies also exhibit superior decorruption capabilities across all levels of severity when compared with DDA~\cite{dda}. Notably, as depicted in Fig.~\ref{fig:severity_granulity} (b), CM shows comparable results with DPM only with 4 NFEs while effectively preserving the object-centric regions of a given image. Note that background colors sometimes change due to the stochastic nature of the diffusion model. 
\begin{figure*}[h!]
    \includegraphics[width=\linewidth]{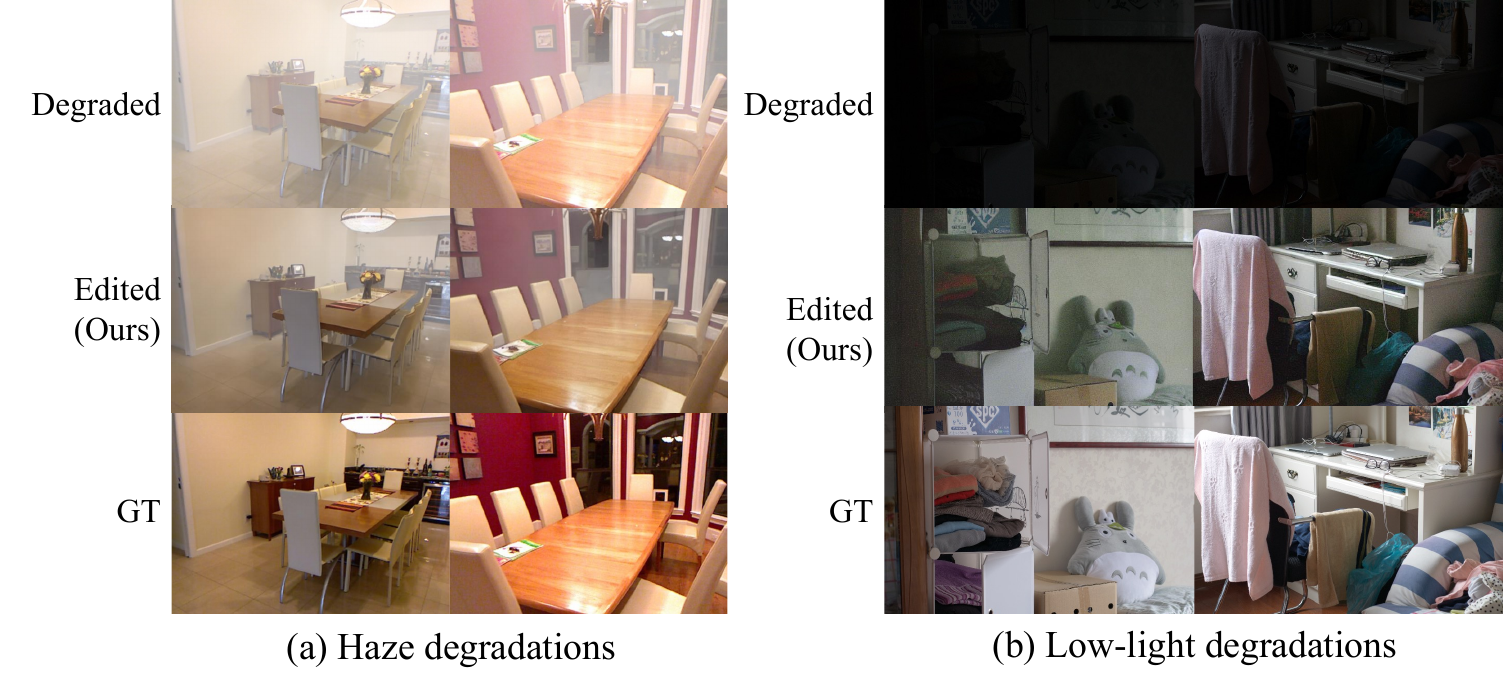} 
    \caption{Further applications of our Decorruptor model in image restoration tasks.}
    \label{fig:applications}
\end{figure*}
\subsection{Further Use-Cases}
Furthermore, as depicted in Fig. \ref{fig:applications}, although such image degradations were not specifically learned in our U-Net fine-tuning stage, our Decorruptor-DPM shows the editing capabilities of corruptions like haze and low-light conditions. The datasets used for these examples are the Reside SOTS~\cite{li2017reside} and LOL~\cite{wei2018deep} datasets, respectively.
\subsection{Additional Results of the Ensemble}
\label{A.5}
\begin{figure}[h!]
    \centering
    \vspace{-0.05in}
    \includegraphics[width=0.9\linewidth]{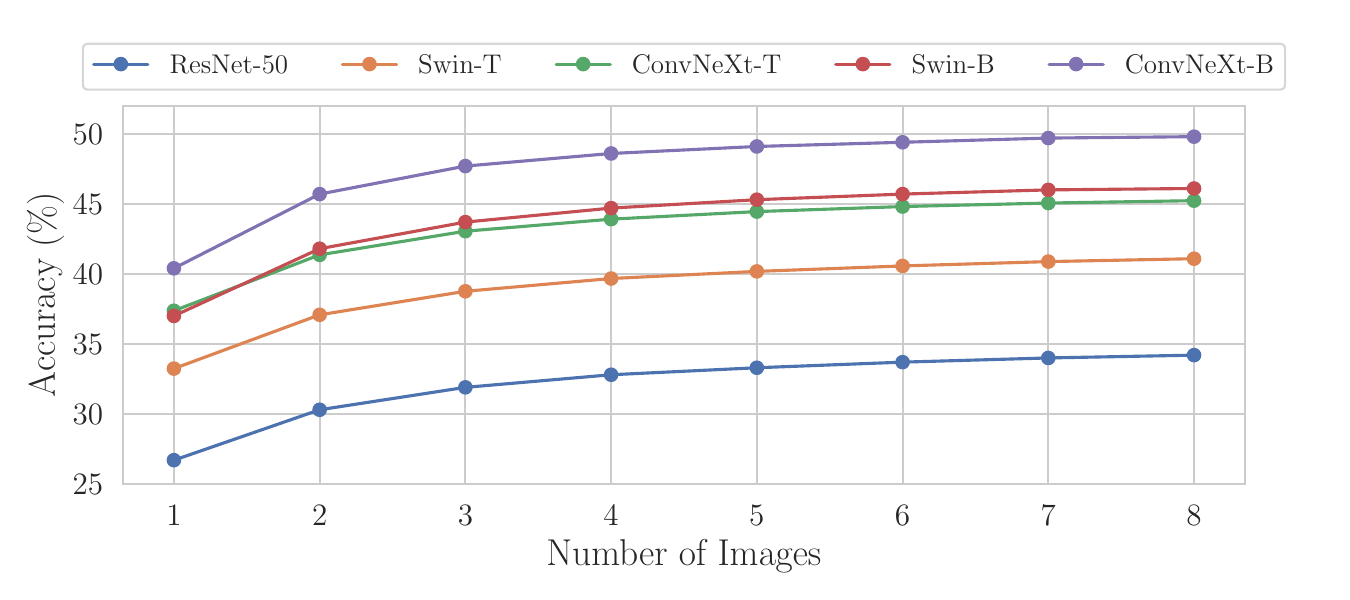}
    \vspace{-0.1in}
    \caption{The accuracy (\%) according to the number of edited images for ensembling in ImageNet-C.}
    \label{fig:fig_ensemble_c}
\end{figure}

\begin{figure}[h!]
    \centering
    \vspace{-0.05in}
    \includegraphics[width=0.9\linewidth]{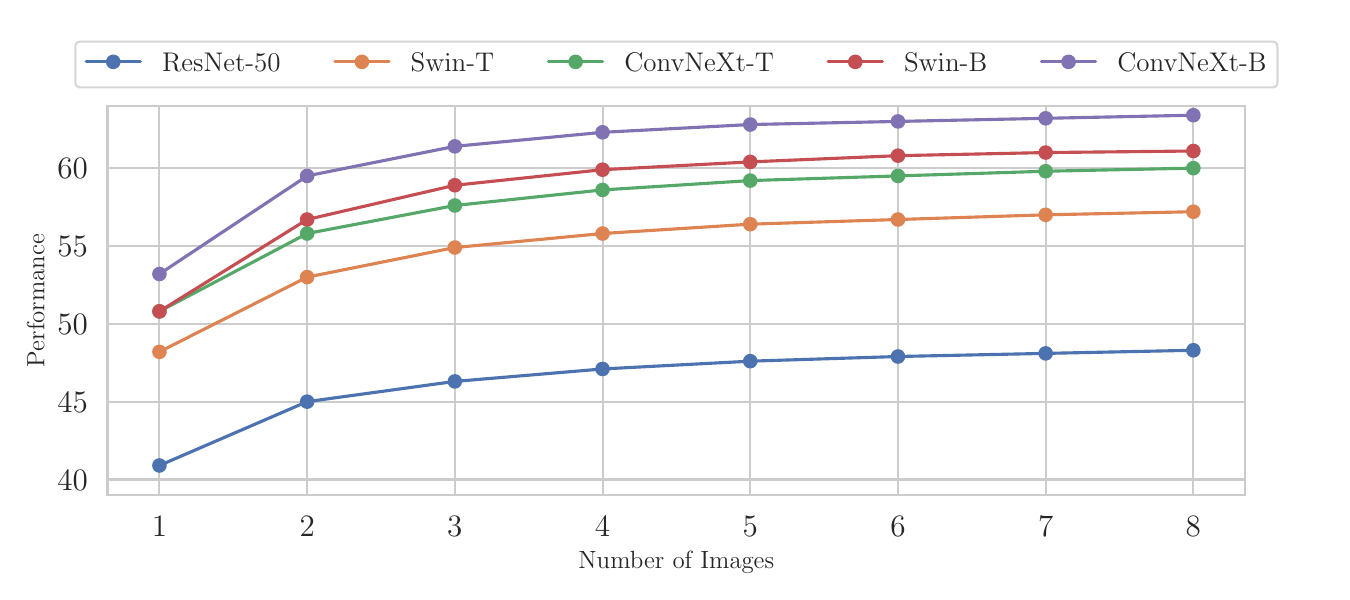}
    \vspace{-0.1in}
    \caption{The accuracy (\%) according to the number of edited images for ensembling in ImageNet-$\bar{\mathrm{C}}$.}
    \label{fig:fig_ensemble_cbar}
\end{figure}

As shown in Table 3 of Section 5.2 in the main text, the addition of an ensemble in Decorruptor-CM led to performance improvement. However, without careful consideration, increasing the number of edited images required for an ensemble results in drawbacks in terms of runtime and memory consumption. Therefore, the number of edited images also becomes a crucial hyperparameter. We illustrated the performance variations with the change in the number of images used for the ensemble in ImageNet-C and ImageNet-$\bar{\mathrm{C}}$ in Figures~\ref{fig:fig_ensemble_c} and~\ref{fig:fig_ensemble_cbar}, respectively. The results indicate a consistent performance increase regardless of the architecture as the number of edited images increases, and the performance tends to converge around 4 ensembles.

\subsection{Diverse Corruption Scenarios, Image Sizes, and Domains}
\begin{figure}[h!]
    \centering    \includegraphics[width=1.0\linewidth]{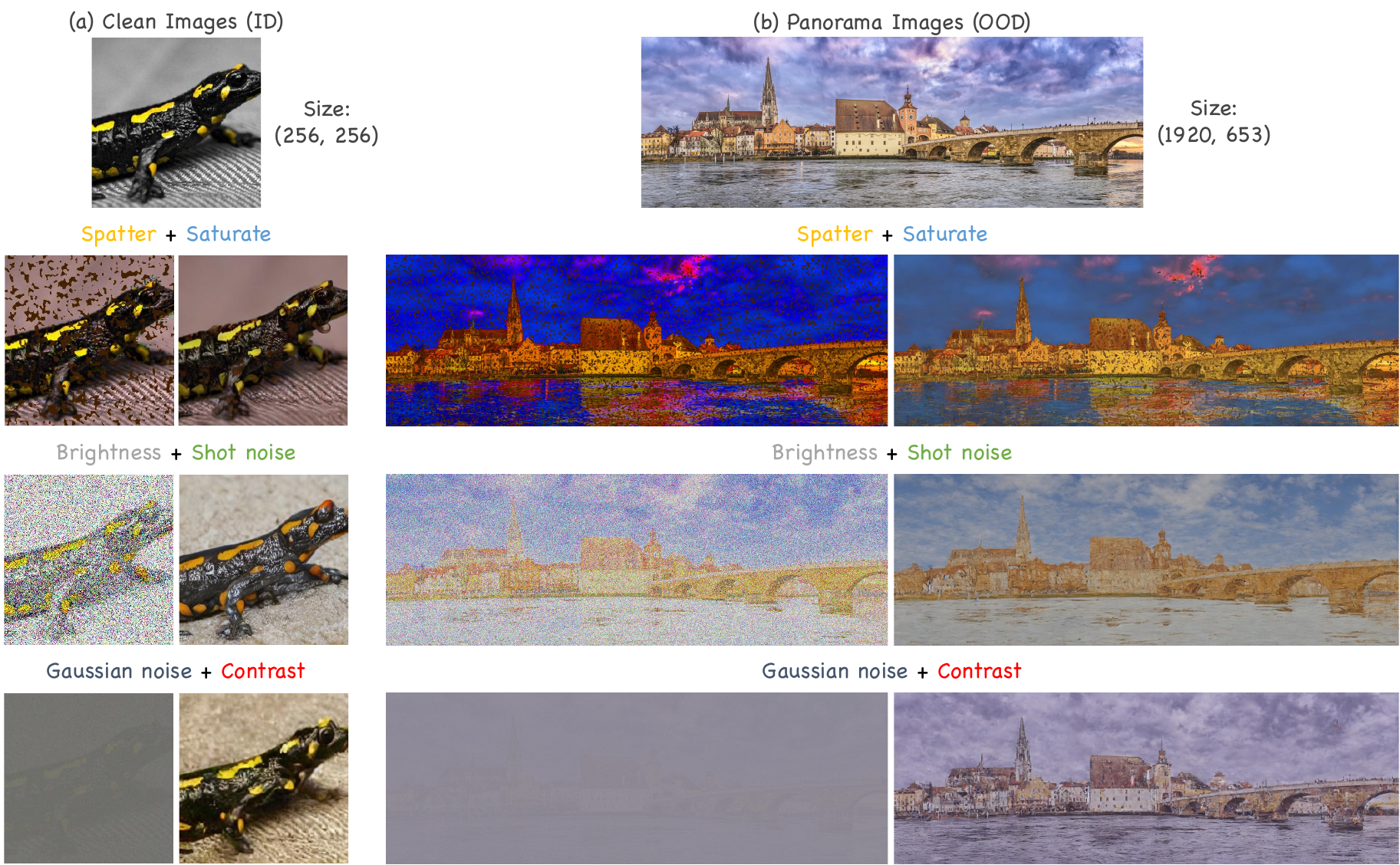}
    \caption{Visualization of experimental results on (a) in-domain and (b) out-of-domain corruption editing performance. We confirmed that our proposed method robustly performs corruption editing even in scenarios with mixed corruption at test time.}
    \label{fig:panoramaa_img}
\end{figure}

In this section, we consider a realistic scenario where an image with various mixed corruptions is encountered at test time. We evaluate the editing performance for this situation using Decorruptor-CM with 4 NFEs. As shown in Fig. \ref{fig:panoramaa_img}, we confirm that corruption editing is feasible in mixed corruption scenarios for both (a) in-domain images and (b) out-of-domain images (\eg, panorama images). We use a mixed corruption severity level of 5 for each type of corruption. In each figure, the left side presents the corrupted images, while the right side displays the edited counterparts.
\section{Ablation Studies}
\subsection{Image Guidance Scaling on Consistency Model}
\begin{figure*}[h!]
    \centering
    \includegraphics[width=\linewidth]{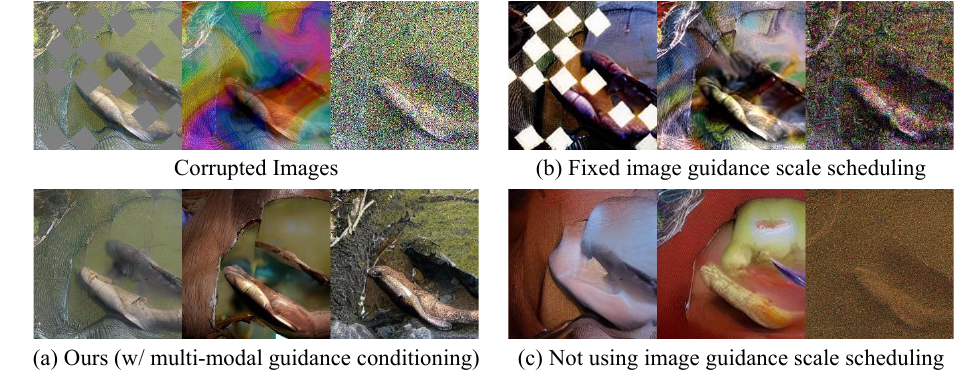} 
    \caption{Ablation studies on (a) using the image guidance scale as conditioning, (b) fixed image guidance scale as 1.3, and (c) not using image guidance scale conditioning during distillation. }
    \label{fig:guidance_scale}
\end{figure*}
For the ablation study, we trained Decorruptor-CM under three conditions: (a) using our multi-modal guidance, (b) using a fixed image guidance, and (c) without using image guidance. Each experiment involved training the model for 12K iterations, consuming 24 GPU hours on an A40 GPU. Fig.~\ref{fig:guidance_scale} highlights the importance of our multi-modal guidance scale. We demonstrate corruption editing performance for checkerboard, Brownian noise, and Gaussian noise. As seen in (a), applying the proposed method by combining it with a text-guidance scale demonstrated the highest performance in corruption editing. In (b), we observe that abnormal images are generated when image guidance scale scheduling is not used for distillation. In (c), editing is minimal when the guidance scale is fixed during distillation, with the images remaining close to the original semantics. Thus, we confirm the importance of a learnable image guidance scale during the distillation stage for effective corruption editing. It is worth noting that guidance scheduling is considered not in CM inference, but only in DPM inference.

\subsection{Using Other Fast Diffusion Schedulers}
\begin{figure*}[h!]
    \centering
    \includegraphics[width=\linewidth]{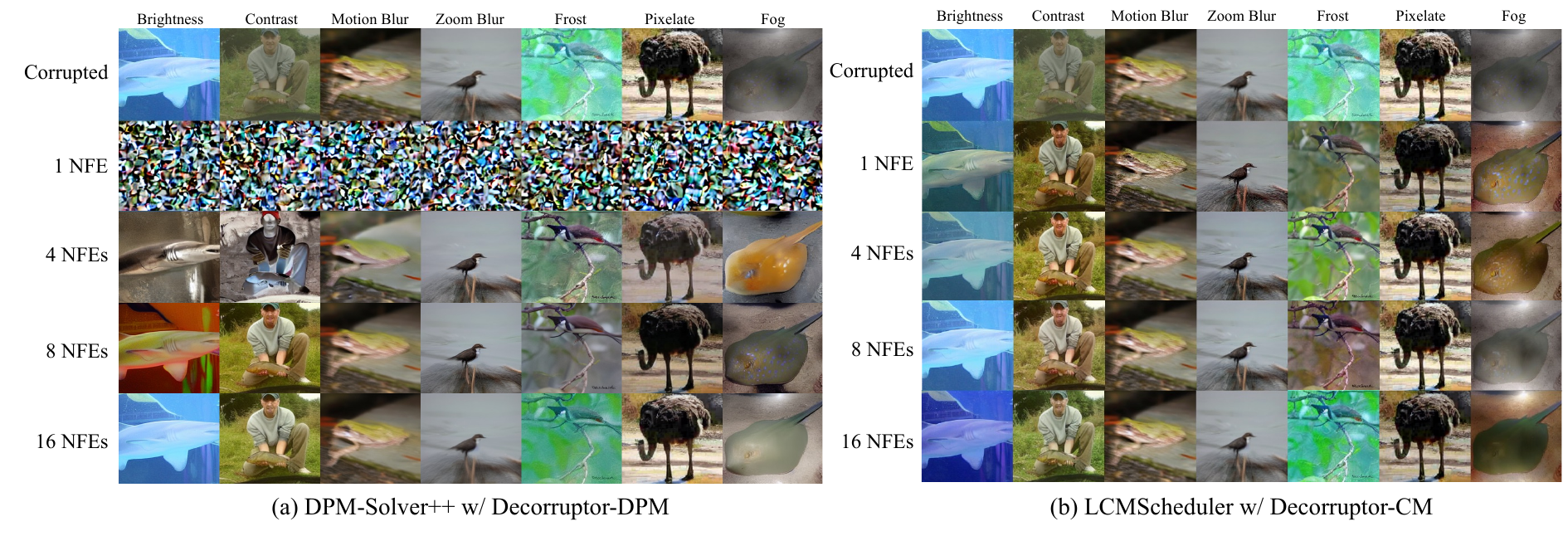} 
    \caption{For several corruptions, (a) a combination of DPM and fast scheduler, (b) results of corruption editing according to the number of NFEs through CM. Note, we used the proposed multi-modal guidance scale conditioning method for the distillation of CM.}
    \label{fig:scheduler}
\end{figure*}
We conducted experiments based on the type of scheduler using the DPM-Solver++ sampler~\cite{lu2022dpm}, traditionally utilized for fast sampling. The results, as shown in Fig.\ref{fig:scheduler} (a), indicate that the sample quality of edited images dramatically decreases with smaller NFEs, with catastrophic failure occurring at 1 NFE. Conversely, as shown in Fig.\ref{fig:scheduler} (b), our Decorruptor-CM demonstrated comparable corruption editing performance at 1 NFE to that at 4 NFEs and showed better editing quality as the NFE increased. This suggests that our proposed Decorruptor-CM enables fast, high-performance corruption editing. Each experiment was conducted with a fixed random seed.

\section{Failure Cases}
\begin{figure*}[h!]
    \centering
    \includegraphics[width=0.8\linewidth]{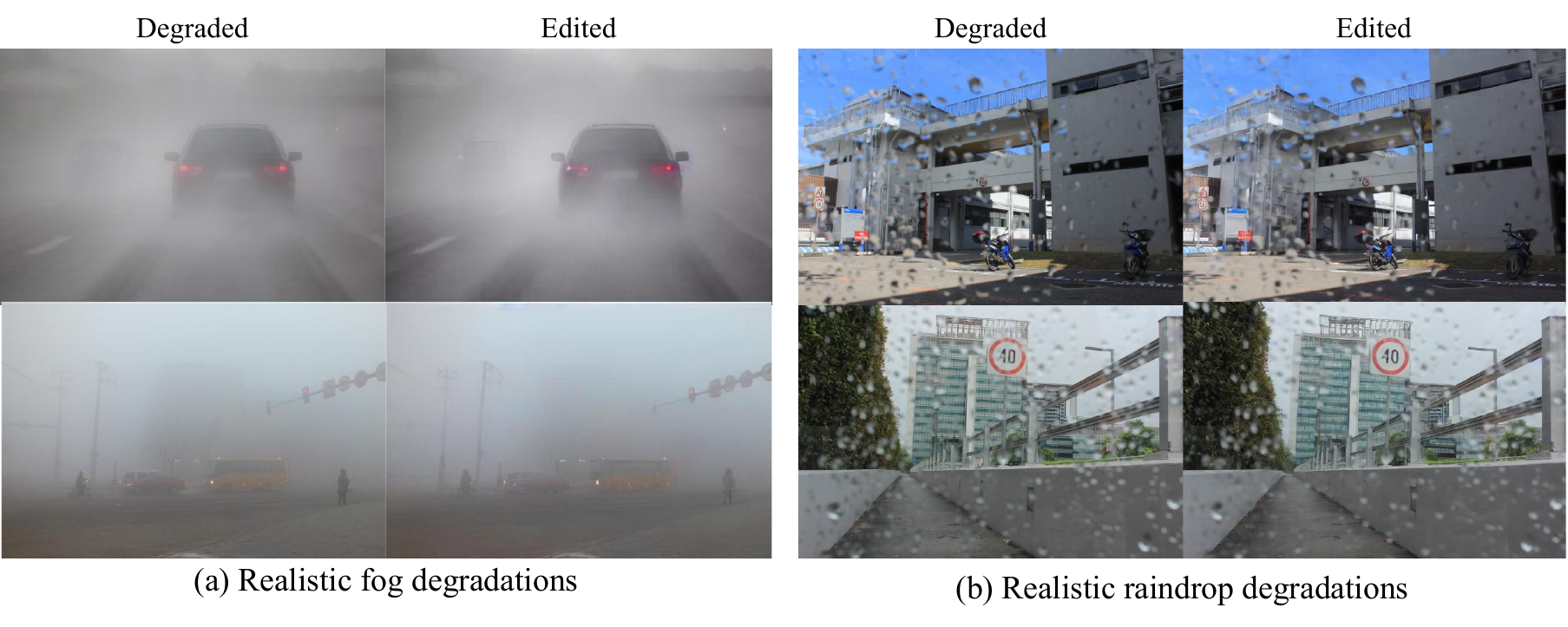} 
    \caption{Failure cases of our model in scenarios involving realistic image degradations.}
    \label{fig:failure_mode}
\end{figure*}
Although our method consistently outperforms other baselines, noticeable improvements were not observed in editing blur and pixelation corruptions. Furthermore, as illustrated in Fig.~\ref{fig:failure_mode}, our model exhibits limitations in corruption editing when faced with more realistic degradations, such as fog or raindrops, which could not be accurately modeled in our corruption modeling scheme. While including paired datasets in the pre-training stage can address such realistic degradations, finding methods to edit such images at test time without having realistic paired images (\ie, clean and corrupted) during training remains a challenging problem for TTA researchers.

\end{document}